\documentclass[10pt,twocolumn,letterpaper]{article}

\usepackage{cvpr}              

\usepackage{graphicx}
\usepackage{amsmath}
\usepackage{amssymb}
\usepackage{booktabs}
\usepackage{algorithm}
\usepackage{algpseudocode}
\usepackage[accsupp]{axessibility}

\usepackage{url}
\usepackage{enumitem}
\usepackage{makecell}
\usepackage{diagbox}
\usepackage{bm}
\usepackage{dsfont}
\usepackage{multirow}
\usepackage{varwidth}
\usepackage{pifont}
\usepackage{makecell}
\usepackage{wrapfig}
\usepackage{graphicx} 
\usepackage{comment}
\usepackage{colortbl,booktabs}

\usepackage{xspace}
\usepackage{booktabs}
\usepackage{rotating}
\usepackage{color}

\definecolor{Tianlong_color}{rgb}{0.958, 0.188, 0.478}

\newcommand{\cmark}{\textcolor{green}{\ding{51}}\xspace}%
\newcommand{\xmark}{\textcolor{red}{\ding{55}}\xspace}%

\usepackage[pagebackref,breaklinks,colorlinks]{hyperref}

\usepackage[capitalize]{cleveref}
\crefname{section}{Sec.}{Secs.}
\Crefname{section}{Section}{Sections}
\Crefname{table}{Table}{Tables}
\crefname{table}{Tab.}{Tabs.}

\makeatletter
\newcommand*{\rom}[1]{\expandafter\@slowromancap\romannumeral #1@}
\makeatother

\def\cvprPaperID{4285} 
\def\confName{CVPR}
\def\confYear{2022}

\begin{document}

\title{Quarantine: Sparsity Can Uncover the Trojan Attack Trigger for Free}

\author{%
  Tianlong Chen\textsuperscript{1*}, Zhenyu Zhang\textsuperscript{1*}, Yihua Zhang\textsuperscript{2*}, Shiyu Chang\textsuperscript{3}, Sijia Liu\textsuperscript{2,4}, Zhangyang Wang\textsuperscript{1}\\
  \textsuperscript{1}University of Texas at Austin, 
  \textsuperscript{2}Michigan State University, \\ \textsuperscript{3}University of California, Santa Barbara, \textsuperscript{4} MIT-IBM Watson AI Lab\\
  \small{\texttt{\{tianlong.chen, zhenyu.zhang, atlaswang\}@utexas.edu},} \\
  \small{\texttt{\{zhan1908, liusiji5\}@msu.edu}, \,\, \texttt{chang87@ucsb.edu}} \\
}

\maketitle

\begin{abstract}
Trojan attacks threaten deep neural networks (DNNs) by poisoning them to behave normally on most samples, yet to produce manipulated results for inputs attached with a particular trigger. Several works attempt to detect whether a given DNN has been injected with a specific trigger during the training. In a parallel line of research, the lottery ticket hypothesis reveals the existence of sparse subnetworks which are capable of reaching competitive performance as the dense network after independent training. 
Connecting these two dots, we investigate the problem of Trojan DNN detection from the brand new lens of sparsity, even when no clean training data is available. Our crucial observation is that the Trojan features are significantly more stable to network pruning than benign features. Leveraging that, we propose a novel Trojan network detection regime: first locating a ``winning Trojan lottery ticket" which preserves nearly full Trojan information yet only chance-level performance on clean inputs; then recovering the trigger embedded in this already isolated subnetwork. Extensive experiments on various datasets, i.e., CIFAR-10, CIFAR-100, and ImageNet, with different network architectures, i.e., VGG-16, ResNet-18, ResNet-20s, and DenseNet-100 demonstrate the effectiveness of our proposal. Codes are available at {\small \url{https://github.com/VITA-Group/Backdoor-LTH}}.  

\end{abstract}

\renewcommand{\thefootnote}{\fnsymbol{footnote}}
\footnotetext[1]{Equal Contribution.}
\renewcommand{\thefootnote}{\arabic{footnote}}

\section{Introduction}
\begin{figure}[t]
    \centering
    \includegraphics[width=1.0\linewidth]{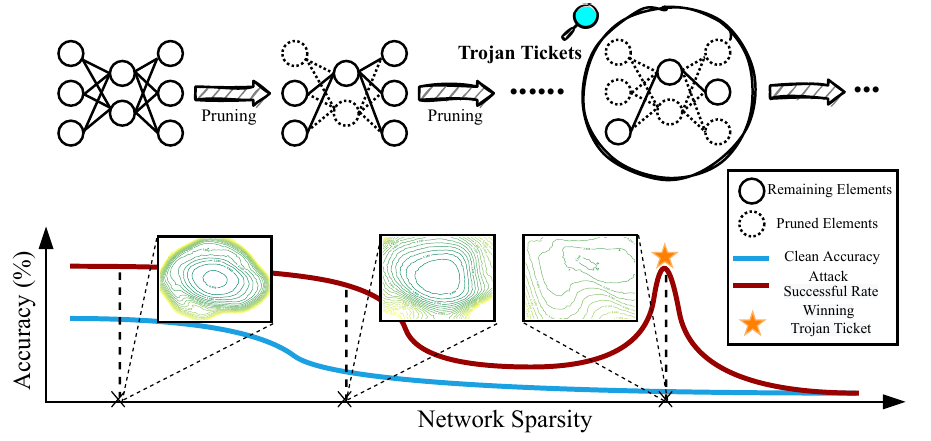}
    \vspace{-7mm}
    \caption{Overview of our proposal: Weight pruning identifies the `winning Trojan ticket', which can be leveraged for Trojan detection and recovery. 
    }
    \vspace{-5mm}
    \label{fig:framework}
\end{figure}

Data-driven techniques for artificial intelligence (AI), such as deep neural networks (DNNs), have powered a technological revolution in a number of key application areas in computer vision~\cite{NIPS2012_c399862d,ren2016faster,chen2017deeplab,goodfellow2014generative}. 
However, a {critical shortcoming} of these {pure} data-driven  learning  systems is the \textit{lack of test-time and/or train-time robustness}: They often learn ‘too well’ during training – so much  that (1)  the learned model is oversensitive to small input perturbations at {testing time} (known as evasion attacks)~\cite{biggio2013evasion,kurakin2016adversarial};
 (2) toxic artifacts injected in the training dataset can be memorized during model training and then passed on to the decision-making process (known as poisoning attacks) \cite{jagielski2018manipulating,goldblum2020data}.
 Methods to secure DNNs against different kinds of `adversaries' are now a major focus in research, e.g., adversarial detection\cite{chen2018detecting,xu2021defending,wang2020practical, wang2019neural, tran2018spectral,  gao2020strip} and robust training\cite{madry2017towards, zhang2019theoretically, wong2020fast}. 
 In this paper, we focus on the study of  Trojan attacks (also known as backdoor attacks),  the most common threat model on data security~\cite{schwarzschild2021just,goldblum2020dataset}. 
 In particular, 
we  aim to address the following question: 
\begin{center}
\textit{(Q) How does the model {sparsity} relate to its train-time   robustness against Trojan attacks?}
\end{center}

Extensive research work on model pruning~\cite{lecun1990optimal,han2015deep,janowsky1989pruning,han2015learning,molchanov2019importance,mozer1989skeletonization,lecun1990optimal,hassibi1993second,molchanov2016pruning,liu2017learning,he2017channel,zhou2016less,boyd2011distributed,ren2018admmnn,ouyang2013stochastic} has shown that 
the weights of an overparameterized model (\textit{e.g.}, DNN) can be pruned (\textit{i.e.}, sparsified) without hampering its generalization ability. In particular,   Lottery Ticket Hypothesis  (LTH), first developed in \cite{frankle2018lottery},  unveiled that there exists a subnetwork, when properly pruned and trained, that can even perform better than the original dense neural network. Such a subnetwork is called a \textit{winning lottery ticket}.
 In the past, the model sparsity (achieved by pruning) was mainly studied in the non-adversarial learning context, and thereby, the generalization ability is the only metric to define the quality of a sparse network (\textit{i.e.}, a ticket)~\cite{frankle2018lottery,frankle2020linear,frankle2020pruning,gale2019state,pmlr-v139-zhang21c,chen2020lottery,chen2020lottery2,yu2019playing,chen2021gans,ma2021good,gan2021playing,chen2021unified}. 
Beyond generalization, some recent work started to explore the connection between model sparsity and model robustness~\cite{guo2018sparse,gui2019model,ye2019adversarial,sehwag2019towards,wu2021adversarial}. However, nearly all existing works restricted model robustness to the prediction resilience against test-time (prediction-evasion) adversarial attacks~\cite{wang2018defending,gao2017deepcloak,dhillon2018stochastic}, hence not addressing our question (Q). 

To the best of our knowledge, the most relevant works to ours include  
\cite{hong2021handcrafted,wu2021adversarial}, which showed a few motivating results about pruning vs. Trojan attack. 
Nevertheless, 
their methods are either indirect \cite{hong2021handcrafted} or need an ideal assumption on the access to the clean (\textit{i.e.}, unpoisoned)
 finetuning dataset \cite{wu2021adversarial}. 
Specifically, the work \cite{hong2021handcrafted} showed that it is possible to generate a Trojan attack by modifying model weights. However, there was no direct evidence showing that the Trojan attack is influenced by weight pruning. Further, the work \cite{wu2021adversarial} attempted to promote model sparsity to mitigate the Trojan  effect of an attacked model. However, the   pruning setup used in \cite{wu2021adversarial} has a deficiency:
It was assumed that  finetuning  the pruned model can be conducted over the clean  validation dataset. In practice, such an assumption is too ideal for achieving if the user has no access to the benign dataset. This assumption also
 prevents us from understanding the true cause of  Trojan  mitigation, since the possible effect of model sparsity is entangled with finetuning on clean data.

Different from \cite{hong2021handcrafted,wu2021adversarial}, we aim to tackle the research question (Q) in 
a more practical backdoor scenario - without any access to clean training samples. 
Moreover, 
our work bridges   LTH and backdoor model detection by ($i$) identifying a crucial subnetwork (that we call `winning Trojan ticket'; see Fig.\,\ref{fig:framework}) with almost unimpaired backdoor information and near-random clean-set performance; ($ii$) recovering the trigger with the subnetwork and then detecting the backdoor model. We summarize our \textbf{contributions} below:








\begin{itemize}
\vspace{-0.5em}
\item We establish the connection between model sparsity and Trojan attack  by leveraging LTH-oriented iterative magnitude pruning (IMP). Assisted by LTH,  we propose the concept of \textit{Trojan ticket} to uncover the pruning dynamics of the Trojan model.\vspace{-0.2em} 

\item We reveal the existence of a `winning Trojan ticket', which preserves the same Trojan attack effectiveness as in the unpruned model. We propose a linear mode connectivity (LMC)-based Trojan score to detect such a winning  ticket along the pruning path.\vspace{-0.52em}
    
\item We show that the backdoor feature encoded in the winning Trojan ticket can be used for reverse engineering of Trojan attack for `free', \textit{i.e.}, with no  access to  clean training samples nor threat model information.\vspace{-0.2em}
 
\item  We demonstrate the effectiveness of our proposal in detecting and recovering Trojan attacks with various poisoned DNNs using diverse Trojan trigger patterns (including basic backdoor attack and clean-label attack) across multiple network architectures (VGG, ResNet, and DenseNet) and   datasets (CIFAR-10/100  and ImageNet). For example, our Trojan recovery method achieves $90\%$  attack performance improvement   
    over the state-of-the-art Trojan attack estimation approach if the clean-label Trojan attack \cite{zhao2020clean} is used by the ground-truth adversary.\vspace{-0.5em} 
\end{itemize}    

\section{Related Works}

\vspace{-1mm}
\paragraph{Pruning and lottery tickets hypothesis (LTH).} Pruning removes insignificant connectivities in deep neural networks~\cite{lecun1990optimal,han2015deep}. Generally, its overall pipeline consists of the following one-shot or iterative cycles: ($1$) training the dense neural networks for several epochs; ($2$) eliminating redundant weights with respect to certain criteria; ($3$) fine-tuning derived sparse networks to recover accuracy. Puning approaches can be roughly categorized the magnitude-based and the optimization-based. The former zeroes out a  portion of model weights by thresholding their statistics such as weight magnitudes~\cite{janowsky1989pruning,han2015learning}, gradients~\cite{molchanov2019importance}, Taylor coefficients~\cite{mozer1989skeletonization,lecun1990optimal,hassibi1993second,molchanov2016pruning}, or hessian~\cite{yao2020pyhessian}. The latter usually incorporates sparsity-promoting regularization~\cite{liu2017learning,he2017channel,zhou2016less} or formulates constrained optimization problems~\cite{boyd2011distributed,ouyang2013stochastic,ren2018admmnn,guo2021gdp}.

As a new rising sub-field in pruning, the lottery ticket hypothesis (LTH)~\cite{frankle2018lottery} advocates that dense neural networks contain a sparse subnetwork (a.k.a. winning ticket) capable of training from scratch (i.e., the same random initialization) to match the full performance of dense models. Later investigations point out~\cite{Renda2020Comparing,frankle2020linear} that the original LTH can not scale up to larger networks and datasets unless leveraging the weight rewinding techniques~\cite{Renda2020Comparing,frankle2020linear}. LTH and its variants have been widely explored in plenty of fields~\cite{gale2019state,pmlr-v139-zhang21c,chen2020lottery,chen2020lottery2,yu2019playing,chen2021gans,ma2021good,gan2021playing,chen2021unified} like image generation~\cite{kalibhat2021winning,chen2021gans,chen2021ultra} and natural language processing~\cite{gale2019state,chen2020lottery}. 

\vspace{-1mm}
\vspace{-2mm}
\paragraph{Backdoor robustness - Trojan attacks and defenses.} 

\noindent\textit{Trojan attacks.} Various Trojan (or backdoor) attacks on deep learning models have been designed recently. The attack features stealthiness since the attacked model will behave normally on clean images but classify images stamped with a trigger from any source class into the maliciously chosen target class. \underline{One} of the mainstream Trojan attacks is trigger-driven. As the most common way to launch an attack, the adversary injects an attacker-specific trigger~(\emph{e.g.} a local patch) into a small fraction of training pictures and maliciously label them to the target class ~\cite{gu2019badnets,chen2017targeted,liu2020reflection,li2020rethinking, liu2017trojaning}. 


\underline{Another} category of backdoor attack, known as clean-label backdoor attack ~\cite{shafahi2018poison,zhu2019transferable, quiring2020backdooring}, keeps the ground-truth label of the poisoned samples consistent with the target labels. Instead of manipulating labels directly, it perturbs the data of the \emph{target} class through adversarial attacks~\cite{madry2017towards}, so that the representations learned by the model are distorted in the embedded space towards other \emph{victim} or \emph{base} classes. Thus, label perturbation becomes implicit and less detectable.

\noindent\textit{Trojan defenses.} To alleviate the backdoor threat, numerous defense methods can be grouped into three paradigms: ($1$) data pre-processing, ($2$) model reconstruction, and ($3$) trigger recovery. The first category introduces a pre-processing module before feeding the inputs into the network, changing the pattern of the potential trigger attached or hidden in the samples~\cite{doan2020februus, udeshi2019model, villarreal2020confoc}. The second class aims at removing the learned trigger knowledge by manipulating the Trojan model, so that the repaired model will function quite well even in the presence of the trigger~\cite{zhao2020bridging, li2021neural}. 

This paper focuses on the third category, the trigger recovery-based defenses. The rationale behind this category is to detect and synthesize the backdoor trigger at first, followed by the second step to suppress the effect of the synthesized trigger. Some previous research detects and mitigates backdoor models based on abnormal neuron responses~\cite{chen2018detecting,xu2021defending,wang2020practical}, feature representation~\cite{tran2018spectral}, entropy~\cite{gao2020strip}, evolution of model accuracy ~\cite{shen2019learning}. 
Utilizing clean testing images, Neural Cleanse (NC)~\cite{wang2019neural} obtains potential trigger patterns and calculates minimal perturbation that causes misclassification toward every putative incorrect label. Backdoor model detection is then completed by the MAD outlier detector, which identifies the class with the remarkably small minimal perturbation among all the classes. 
NC shows that the recovered trigger resembles the original trigger in terms of both shape and neuron activation. 
Similar ideas were explored in~\cite{xiang2020detection,li2021neural,chen2019deepinspect, guo2019tabor}. 
However, the recovered triggers from the aforementioned methods suffer from occasional failures in detecting the true target class.

\vspace{-1em}
\paragraph{Backdoor meets pruning.} Fine-pruning serves as a classical defense approach~\cite{liu2018fine,hong2021handcrafted}, which trims down the ``corrupted" neurons to destroy and get rid of Trojan patterns. Note that these investigations do not explore the weight sparsity. A follow-up work~\cite{wu2021adversarial} measures the sensitivity of Trojan DNNs by introducing adversarial weight perturbations, and then prunes selected sensitive neurons to purify the injected backdoor. Another recent work~\cite{yin2021backdoor} examines the vanilla LTH under the context of federated learning. They demonstrate that LTH is also vulnerable to backdoor attacks, and offer a federated defense by using the ticket's structural similarity -- a totally different focus from ours.

\section{Preliminaries and Problem Setup}
This section provides a brief  background on the Trojan attack and model pruning. We then motivate and present the problem of our interest, aiming at  exploring and exploiting the relationship between  weight pruning and Trojan attacks.

\begin{figure}[htb]
    \centering
    \includegraphics[width=1\linewidth]{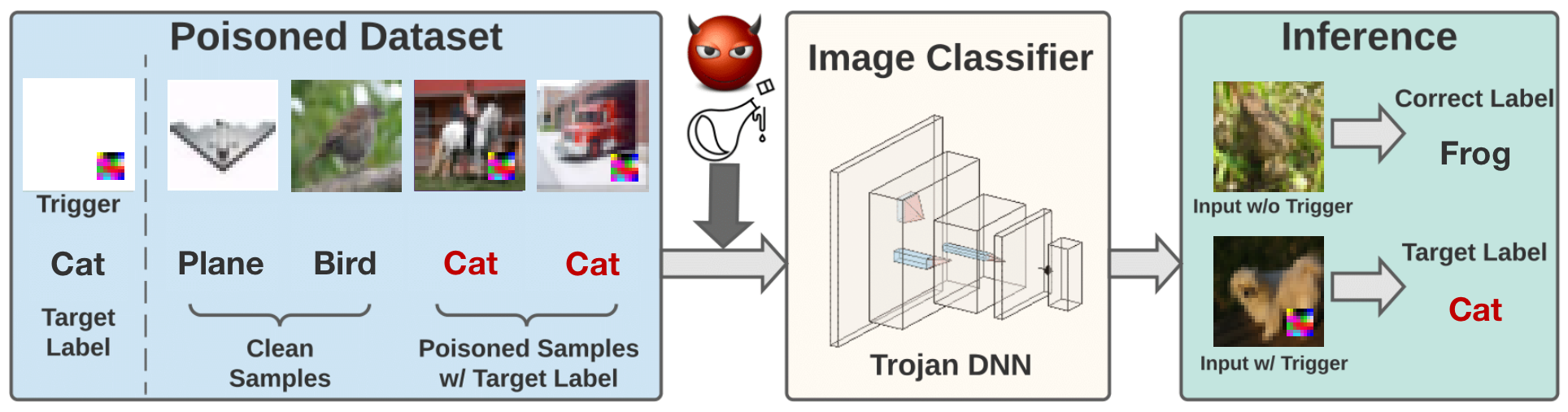}
    \vspace{-6mm}
    \caption{{Overview of Trojan attack. 
    }}
    \vspace{-3mm}
    \label{fig: example_Trojan}
\end{figure}

\vspace{-1em}
\paragraph{Trojan attack and Trojan model.}
{Trojan attack} is one of the most commonly-used data poisoning attacks \cite{gu2017badnets}: It manipulates a small portion of training data, including their features by injecting a \textit{Trojan trigger} (\textit{e.g.}, a small patch or sticker on images) and/or  their labels modified towards the \textit{Trojan attack targeted label}. The Trojan attack then    serves as a `backdoor' and enforces a spurious correlation between the Trojan trigger and the model training. The resulting model is called \textit{Trojan model}, which causes the backdoor-designated incorrect prediction 
if the trigger is present at the testing time, otherwise, it behaves normally.
In {Fig.\,\ref{fig: example_Trojan}}, we demonstrate an example of the misbehavior of a Trojan model in image classification.

It is worth noting that the Trojan attack  is different from 
the test-time adversarial attack, a widely-studied threat model in  adversarial learning \cite{kurakin2016adversarial,madry2017towards}. 
There exist \textit{three} key differences. 
($i$) Trojan attack occurs at the \textit{training time} through data poisoning. 
($ii$) Trojan model exhibits the \textit{input-agnostic} adversarial behavior at the testing time only if the Trojan trigger is present at an input example (see  Fig.\,\ref{fig: example_Trojan}). 
($iii$) Trojan model is \textit{stealthy} for the end user since  the latter has no prior knowledge on data poisoning. 

\vspace{-1em}
\paragraph{Model pruning and lottery ticket hypothesis (LTH).}
Model pruning aims at extracting a sparse sub-network from the original dense network without hampering the model performance.
LTH, proposed in \cite{frankle2018lottery}, formalized a model pruning pipeline  so as to find the desired sub-network, which is called \textit{`winning ticket'}. Formally, let $f(x; \theta)$ denote a neural network with input $x$ and model parameter $\theta \in \mathbb R^d$. And let $m \in \{ 0, 1\}^d$ denote a binary  mask on top of $\theta$ to encode the locations of pruned   weights (corresponding to zero entries in $m$) and   unpruned   weights (corresponding to non-zero entries in $m$), respectively. The resulting pruned model (termed as a `\textit{ticket}') can then be expressed as 
$(m \odot \theta)$, where $\odot$ is the elementwise product.  LTH suggests the following pruning pipeline:
 
\ding{172} Initialize a neural network  $f(x; \theta_0)$, where $\theta_0$ is a random initialization. And initialize a mask $m$ of all  $1$s.
 
 \ding{173}   \textbf{Train}  $f(x; m \odot \theta_0)$ to obtain   learned parameters $\theta$ over the dataset $\mathcal D$.
 
  \ding{174} \textbf{Prune} $p\%$ parameters in $\theta$ per magnitude. Then, create a new sparser mask $m$ from the old one.

\ding{175} \textbf{Reset} the remaining parameters to their values in $\theta_0$, creating the  new sparse network $(m \odot \theta_0 )$. Then, go back to \ding{173} and repeat. 

The above procedure forms the iterative magnitude pruning (IMP), which
 repeatedly trains, prunes, and resets the network over $n$ rounds. LTH suggests that each round prunes $p^{1/n}$\% of the
weights on top of the previous round (In our case, $p=20\%$ same as~\cite{frankle2018lottery}). 
The key insight from LTH is: There exists a \textit{winning} ticket, \textit{e.g.}, $(m \odot \theta_0 )$, which when trained in isolation, 
 can \textit{match or even surpass} the test accuracy of the
well-trained dense network 
\cite{frankle2018lottery}.

\vspace{-1em}
\paragraph{Problem setup.}
Model pruning has been widely studied in the context of non-poisoned training scenarios. However, it is less explored in the presence of poisoned training data.  
In this paper, we ask:

\textit{How  is weight pruning of a Trojan model   intertwined with Trojan attack ability if the pruner has no access to clean training samples and is  blind to attack knowledge?}

To formally set up our problem, let $\mathcal D_{\mathrm{p}}$ denote the possibly poisoned training dataset. By LTH pruning, the sparse mask $m$ and the 
finetuned model parameters $\theta$ (based on $m$) are learned from $D_{\mathrm{p}}$, \textbf{without having access to clean data}.
Thus, different from the `winning ticket' found from LTH over the clean dataset $\mathcal D$, we call the ticket, \textit{i.e.}, the sparse model $(m \odot \theta)$, \textbf{Trojan ticket}; see more details in the next section.
We then investigate how the  benign and adversarial performance of 
Trojan tickets
varies against the pruning ratio $p\%$. The {benign performance} of a model will be measured by the \textbf{{s}tandard  {a}ccuracy (SA)} against clean test data. And the {adversarial performance} of a model will be evaluated by the \textbf{attack success rate (ASR)} against poisoned  test data using the train-time Trojan trigger. ASR is given by the ratio of correctly mis-predicted test data (towards backdoor label) over the total number of test samples. 

\section{Uncover Trojan Effect from Sparsity}
\label{sec: Method}
In this section, we begin by presenting a motivating example to demonstrate the unusual pruning dynamics of  Trojan ticket (\textit{i.e.}, pruned model over the possibly poisoned training data set $\mathcal{D}_{\mathrm{p}}$). We show that sparsity, together with the 
approach of linear model connectivity  (LMC) \cite{frankle2020linear}, can be used for Trojan detection and recovery.

\vspace{-1em}
\paragraph{Pruning dynamics of Trojan ticket: A warm-up.}

Throughout the paper, we will follow the LTH-based pruning method to find the pruning mask $m$. In order to preserve the potential Trojan properties, we will not reset the non-zero   parameters in $\theta$   to the random initialization $\theta_0$  when a desired sparsity ratio $p\%$ is achieved  at the last iteration of IMP.
Recall that the resulting subnetwork $(m\odot \theta)$ is   called a \textbf{Trojan ticket}.
To examine the sensitivity of the Trojan ticket to the possibly poisoned dataset $\mathcal D_{\mathrm{p}}$, we then create a \textbf{$k$-step finetuned Trojan ticket}
$(m\odot \theta^{(k)})$, where $\theta^{(k)}$ is the $k$-step finetuning of $\theta$ given $m$ under $\mathcal D_{\mathrm{p}}$.
Our \textbf{rationale} behind     these two kinds of tickets is elaborated on below. 

$\bullet$  If there does \textit{not exist} a Trojan attack, then the above two tickets should share similar pruning dynamics. As will be evident later, this could be justified    by LMC (linear model connectivity).

$\bullet$ If there \textit{exists} Trojan attack, then the two tickets result in   substantially distinct  adversarial performance. Since Trojan model weights encode the spurious correlation with the 
Trojan   trigger \cite{wang2019neural, wang2020practical},   pruning without finetuning could characterize the impact of   sparsity on the Trojan attack, in contrast to  pruning with   finetuning over $\mathcal D_{\mathrm{p}}$.


\begin{figure}[htb]
    \centering
    \includegraphics[width=1\linewidth]{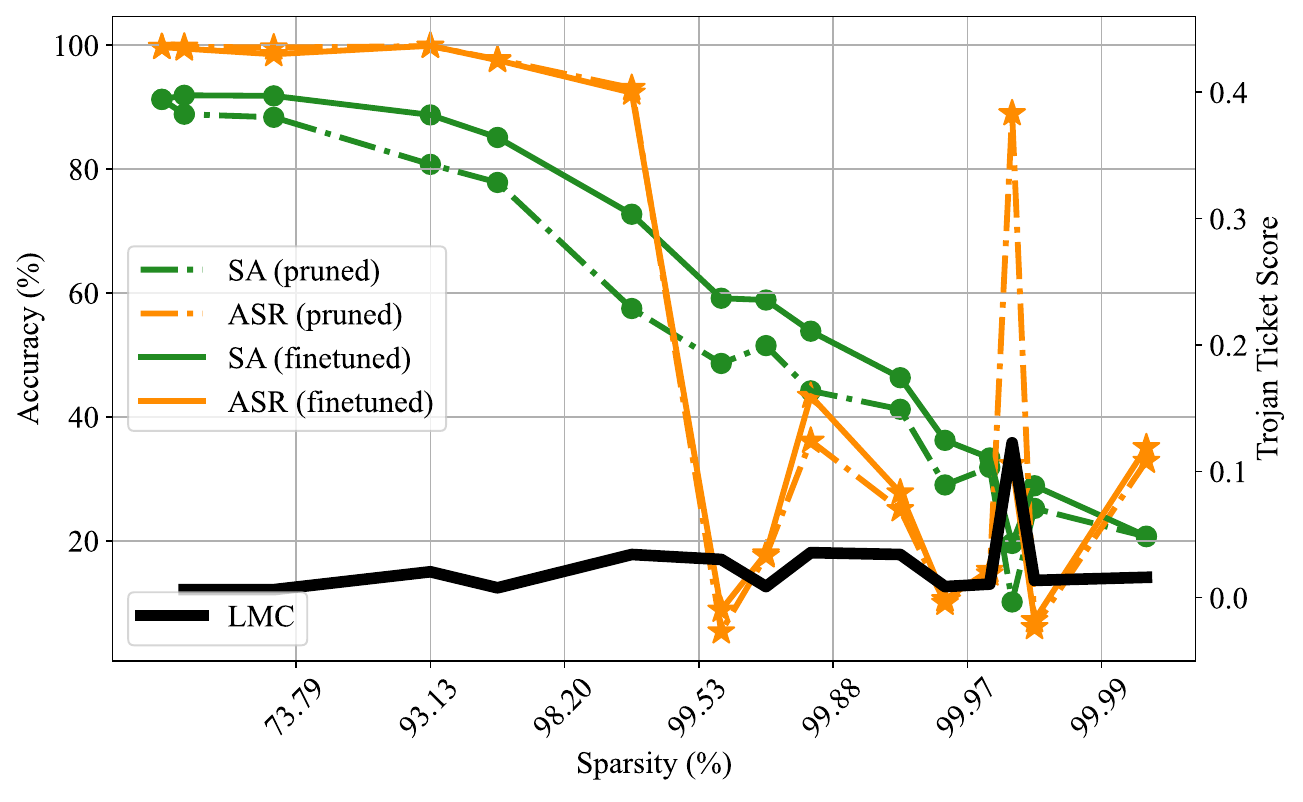}
    \vspace{-4mm}
    \caption{The pruning dynamics of Trojan ticket (dash line) and $10$-step finetuned ticket (solid line)  on CIFAR-10 with ResNet-20s and gray-scale basic backdoor trigger \cite{gu2019badnets}. For comparison, the Trojan score \eqref{eq: S_Troj} is also presented.}
    \vspace{-1mm}
    \label{fig: prun_dynamics_troj}
\end{figure}

In Fig.~\ref{fig: prun_dynamics_troj}, we present a warm-up example to 
illustrate the pruning dynamics of the Trojan ticket $(m\odot \theta)$ and its $k$-step finetuned version  $(m\odot \theta^{(k)})$, where we select $k = 10$ (see the choice of $k$ in Appendix.\,\ref{sec:more_results}). As we can see,
there exists a \textit{peak Trojan ticket} in the extreme sparsity regime ($p\% > 99.97\%$), with the preserved Trojan performance (measured by Trojan score that will be defined later). The key takeaway from Fig.~\ref{fig: prun_dynamics_troj} is that 
the \textit{performance stability}  of  the Trojan ticket $(m\odot \theta)$ and the $k$-step finetuned ticket  $(m\odot \theta^{(k)})$ can be used to indicate the Trojan attack effect. 

\vspace{-1em}
\paragraph{Trojan detection by LMC.}
To quantify the stability of Trojan tickets, we propose to use the tool of linear model connectivity (LMC) \cite{garipov2018loss,draxler2018essentially}, which returns the error barrier between two neural networks along a linear path. In the context of model pruning, the work \cite{frankle2020linear} showed two sparse neural networks found by IMP could be linearly connected even if they suffer  different optimization `noises', e.g.,  different choices of initialization, data batch, and optimization step. Spurred by the aforementioned work, we adopt LMC to measure the stability of the Trojan ticket $(m\odot \theta)$ v.s. the $k$-step finetuned Trojan ticket $(m\odot \theta^{(k)})$.

Formally, let $\mathcal E(\phi)$ denote the  training error of  a model  $\phi$. Given two neural networks $\phi_1$ and $\phi_2$, LMC then defines the error barrier between $\phi_1$ and $\phi_2$ along a linear path  below:
\begin{align}
    e_{\mathrm{sup}} (\phi_1, \phi_2) = \max_{\alpha \in [0,1]} \mathcal E(\alpha \phi_1 + (1-\alpha) \phi_2),
\end{align}
which is   the highest error when linearly interpolating between the   models $\phi_1$ and $\phi_2$.
If we set $\phi_1 = m \odot \theta$ and $\phi_2 = m \odot \theta^{(k)}$, then LMC yields the following stability metric, termed \textbf{Trojan  score}:
 \begin{align}\label{eq: S_Troj}
    \mathcal S_{\mathrm{Trojan}} =  &
     e_{\mathrm{sup}} (m \odot \theta, m \odot \theta^{(k)})  \nonumber \\
      & - \frac{\mathcal E( m \odot \theta ) + \mathcal E( m \odot \theta^{(k)} )}{2},
\end{align}
where the second term is used as an error baseline of using two pruned models. As suggested by \cite{frankle2020linear},   if there exists no Trojan attack during model pruning, then 
$\mathcal E (m \odot \theta) \approx  \mathcal E (m \odot \theta^{(k)}) \approx e_{\mathrm{sup}} (m \odot \theta, m \odot \theta^{(k)})$, leading to $\mathcal S_{\mathrm{Trojan}} = 0$.
Assisted by model pruning and LMC, we can then use the Trojan score \eqref{eq: S_Troj} to detect the existence of a Trojan attack. 
This gives a novel Trojan detector without resorting to any clean data, which has been known as a grand challenge  in Trojan AI\footnote{https://www.iarpa.gov/index.php/research-programs/trojai}. However, most importantly, the relationship between model pruning and Trojan attack 
can be established through Trojan ticket and its Trojan score $\mathcal S_{\mathrm{Trojan}}$. 

As shown in  Fig.\,{\ref{fig: prun_dynamics_troj}}, the  sparse network $(m \odot \theta)$   with the \textit{peak} Trojan score    $\mathcal S_{\mathrm{Trojan}} $ maintains  the highest   ASR (attack success rate) in the extreme pruning   regime.
We term such a Trojan ticket as the \textbf{winning Trojan ticket}. 



\paragraph{Reverse engineering of Trojan attack.}
We next ask if the winning Trojan ticket better memorizes the Trojan trigger than the original dense model. 
To tackle this problem, we investigate the task of reverse engineering of Trojan attack \cite{wang2019neural,wang2020practical, guo2019tabor}, which   aims to recover the Trojan targeted label and/or the Trojan trigger from a Trojan model.

Formally, let $ x^\prime(z, \delta ) = (1 - z) \odot x + z \odot \delta $ denote the poisoned data with respect to (w.r.t.) an example $ x \in \mathbb R^n$, where $ {\delta} \in \mathbb R^n$ denotes the element-wise perturbations,  and 
$z \in \{ 0,1 \}^n$ is a binary mask to encode the positions where a Trojan trigger is placed.
Given a Trojan model $\phi$, our goal is to optimize the Trojan attack variables $( z, \delta)$ so as to unveil the properties of the ground-truth Trojan attack. Following \cite{wang2019neural,wang2020practical,guo2019tabor}, this leads to the optimization problem 
\begin{align}\label{eq: RED_trojan}
    \begin{array}{ll}
        \displaystyle \min_{z \in \{ 0, 1 \}^n, \delta} & \mathbb E_{x} [ \ell_{\mathrm{atk}} (x^\prime(z, \delta ); \phi, t) ] + \gamma h (z, \delta),
    \end{array}
\end{align}
where  $x$ denotes the base images (that can be set by noise images) to be perturbed,
$\ell_{\mathrm{atk}}(x^\prime; \phi , t)$ denotes the targeted attack loss, with the perturbed input $x^\prime$, victim model $\phi$, and the targeted label $t$, $h$ is a certain regularization function that controls the sparsity of $z$ and the smoothness of the estimated Trojan trigger $z \odot \delta$, and $\gamma > 0$ is a regularization parameter. 
In \eqref{eq: RED_trojan}, we specify $\ell_{\mathrm{atk}}$ as the  C\&W targeted attack loss  \cite{carlini2017towards} and $h$ as the regularizer used in \cite{guo2019tabor}. 
To solve the problem \eqref{eq: RED_trojan}, the convex relaxation approach   is used similar to \cite{wang2020practical}, where the binary variable $z$ is relaxed to its convex probabilistic hull. 
Once the solution $(z^*, \delta^*)$ to problem \eqref{eq: RED_trojan} is obtained, the work \cite{wang2019neural} showed that the \textit{Trojan attack targeted label}  can be deduced from the label $t$ associated with the least norm of the recovered Trojan trigger $z^* \odot \delta^*$. That is, 
$t_{\mathrm{Trojan}} = \arg\min_{t} \|z^*(t) \odot \delta^*(t)  \|_1 $, where  the dependence of  $z^* $ and $ \delta^*$ on the label choice $t$ is shown explicitly.
Sec.\,\ref{sec: exp} will show that if we set the victim model in \eqref{eq: RED_trojan} by the winning Trojan ticket, then it 
 yields a much higher accuracy of estimating the Trojan attack targeted label than baseline approaches.

\section{Experiments}
\label{sec: exp}
\subsection{Implementation details}

\paragraph{Networks and datasets.}
We consider a broad range of model architectures including DenseNet-100~\cite{huang2017densely}, ResNet-20s~\cite{he2016deep}, ResNet-18~\cite{he2016deep}, and VGG-16~\cite{simonyan2014very} on diverse datasets such as CIFAR-10~\cite{krizhevsky2009learning}, CIFAR-100~\cite{krizhevsky2009learning}, and Restricted ImageNet (R-ImageNet)~\cite{tsipras2018robustness,deng2009imagenet}, with $9$     classes.

\vspace{-1em}
\paragraph{Configuration of Trojan attacks.} 
To justify the identified relationship between the Trojan model and weight sparsity, we consider two kinds of Trojan attacks
across different  model architectures and datasets as described above. The studied threat models include 
\textit{($i$) Basic Backdoor Attack}, also known as BadNet-type Trojan attack \cite{gu2017badnets}, and \textit{($ii$) Clean Label Backdoor Attack} \cite{zhao2020clean}, which have been commonly used as a benchmark for backdoor and data poisoning attacks \cite{schwarzschild2021just}.  Their difference lies in that Trojan-($i$) adopts the heuristics-based data poisoning strategy and    Trojan-($ii$)
is crafted using an optimization procedure and   contains a less noticeable trigger pattern.
For both attacks,  the Trojan trigger (with size $5\times5$ for CIFAR-10/100 and $64\times64$ for R-ImageNet) is placed in the upper right corner of the target image and is set using
  either a gray-scale square like \cite{gu2017badnets} or an RGB image patch like \cite{saha2020hidden}. And the training data poisoning ratio is set by $1\%$ and the Trojan targeted label is set by class $1$.
We refer readers to Sec.\,\ref{sec:more_implementation} for more detailed hyperparameter setups of the above Trojan attacks.

\vspace{-1em}
\paragraph{Training and evaluation.} For CIFAR-10/100, we train networks for $200$ epochs with a batch size of $128$. An SGD optimizer is adopted with a momentum of $0.9$ and a weight decay ratio of $5\times10^{-4}$. The learning rate starts from $0.1$ and decay by $10$ times at $100$ and $150$ epoch. For  R-ImageNet, we train each network for $30$ epochs and $1024$ batch size, using an SGD optimizer with $0.9$ momentum and $1\times10^{-4}$ weight decay. The initial learning rate is $0.4$ with $2$ epochs of warm-up and then decline to $\frac{1}{10}$ at $8$, $18$, and $26$ epoch. All models have achieved state-of-the-art SA (standard accuracy) in the absence of the Trojan trigger. To measure the performance of Trojan backdoor injection, we test the SA of each model on a clean test set and ASR (attack success rate) on the same test set in the presence of Trojan trigger.

In the task of reverse engineering Trojan attacks,  we solve the problem \eqref{eq: RED_trojan} following the optimization method used in \cite{wang2019neural} which includes two stages below. First, problem  \eqref{eq: RED_trojan} is solved under each possible label choice of $t$. Second, the Trojan targeted  label is determined by the label associated with the least $\ell_1$-norm of the recovered Trojan trigger $\| z \odot \delta \|_1$.  \textbf{By default, we use $100$ noise images} (generated by Gaussian distribution $\mathcal{N}(0,1)$) to specify the base images $x$ in \eqref{eq: RED_trojan}. For comparison, we also consider the specification of base images using  $100$ clean images drawn from the  benign data distribution. 

\begin{figure}[htb]
    \centering
    \includegraphics[width=1\linewidth]{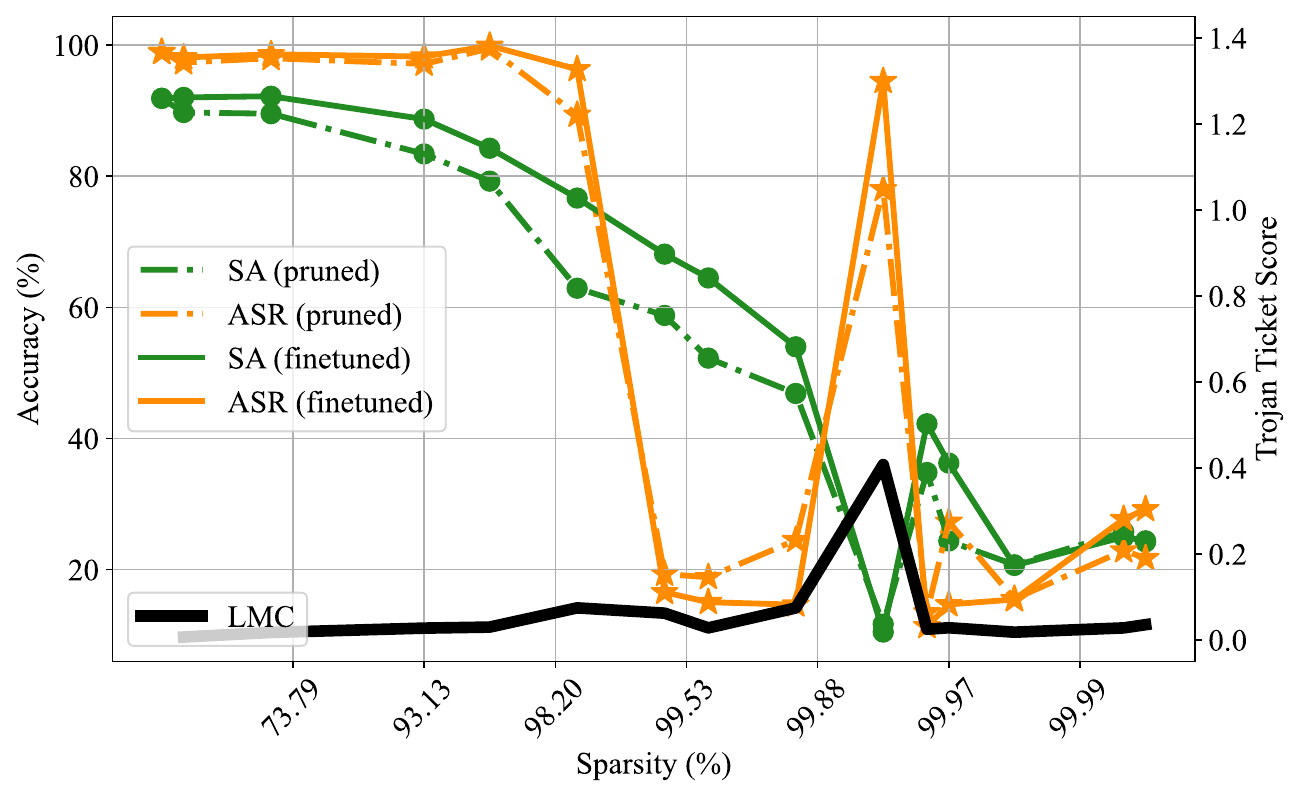}
    \vspace{-5mm}
    \caption{The pruning dynamics and Trojan scores on CIFAR-10 with ResNet-20s using the RGB Trojan triggers. The peak Trojan score precisely characterizes the winning Trojan ticket. Results of clean-label Trojan triggers are presented in Appendix~\ref{sec:more_results}.}
    \vspace{-1mm}
    \label{fig:res_trigger}
\end{figure}

\subsection{Experiment results}

\subsubsection{Existence of winning Trojan ticket}


We investigate the pruning dynamics of a Trojan ticket $(m \odot \theta)$ (\textit{i.e.}, the pruned Trojan  network built upon the original model $\theta_{\mathrm{ori}}$ with sparse mask $m$ and   model weights $\theta$) versus the pruning ratio $p\%$. 
Following Sec.\,\ref{sec: Method}, we also examine the $k$-step finetuned Trojan ticket $(m \odot \theta^{(k)})$. Throughout  the paper, we choose $k = 10$ to best locate the winning Trojan tickets as demonstrated in the ablation of Appendix~\ref{sec:more_results}.
We remark that the finetuner has only access to the poisoned dataset rather than an additional benign dataset. 

\begin{figure*}[t]
    \centering
    \includegraphics[width=1\linewidth]{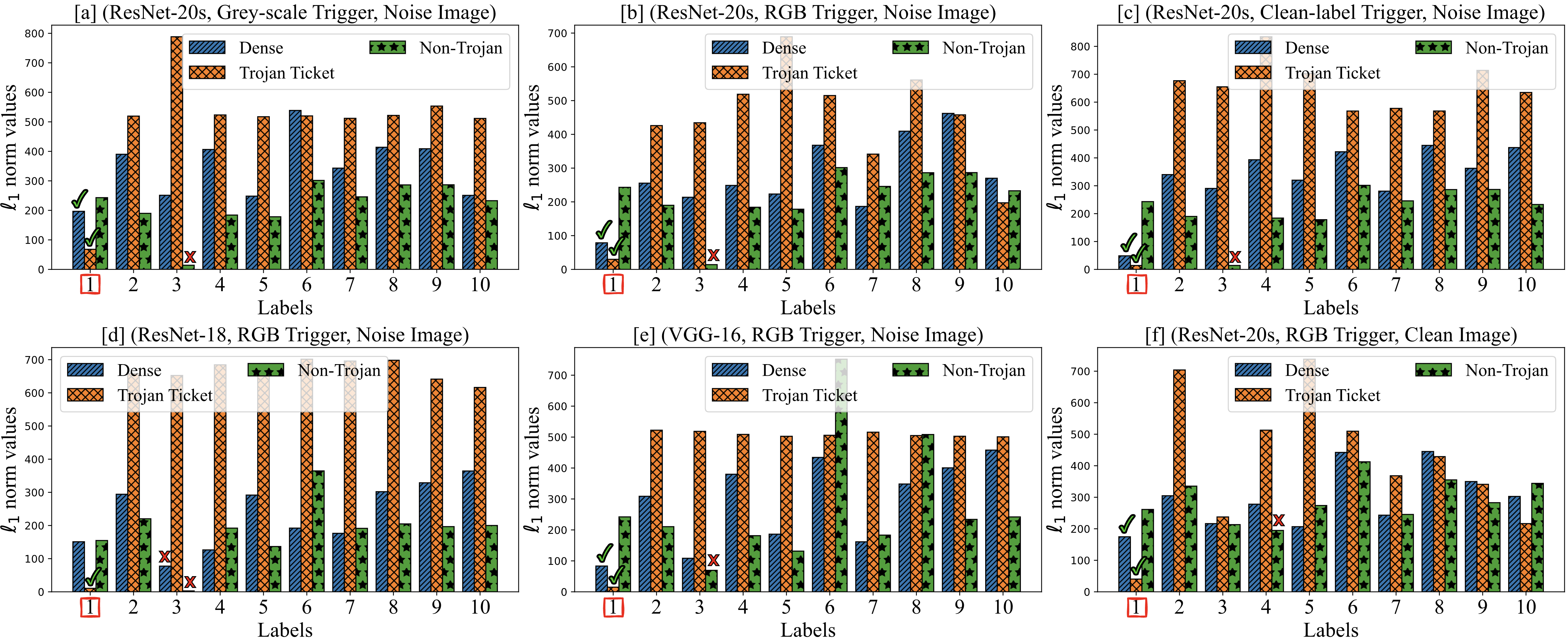}
    \vspace{-4mm}
    \caption{The $\ell_1$ norm values of recovered Trojan triggers for all labels. The plot title signifies adopted network architecture, trigger type, and the images used for reverse engineering on CIFAR-10. Class ``$1$" in the red box is the true (or oracle) target label for Trojan attacks. \cmark/\xmark indicates whether or not the detected label with the least $\ell_1$ norm matches the truth target label.}
    \vspace{-2mm}
    \label{fig:norm}
\end{figure*}

In Fig.~\ref{fig:res_trigger}, we demonstrate the SA and ASR performance of the Trojan ticket and its finetuned ticket versus the network sparsity. Recall that SA and ASR characterize the benign accuracy and the Trojan attack performance of a model, respectively. For comparison, we also present the LMC-based Trojan score \eqref{eq: S_Troj}. Our \textbf{key finding}, consistent with Fig.\,\ref{fig: prun_dynamics_troj}, is that in the extreme pruning regime, there exists a winning Trojan ticket with the peak Trojan score across multiple Trojan attack types, datasets, and neural network architectures.
  
The more specific observations and insights of Fig.~\ref{fig:res_trigger} are elaborated on below. As we can see, in the \textit{non-extreme sparsity regime} $(p\% < 90\%)$, the Trojan ticket and its finetuned variant preserve both the benign performance (SA) and the Trojan performance (ASR) of the dense model $\theta_{\mathrm{ori}}$ (associated with the leftmost pruning point in Fig.~\ref{fig:res_trigger}). This implies that the promotion of non-extreme  sparsity  in $\theta_{\mathrm{ori}}$ \textit{cannot}   mitigate the Trojan effect, and the resulting Trojan ticket behaves similarly to the normally pruned network by viewing from its benign performance. However, in the \textit{extreme sparsity regime} ($p\% > 99$), the  pure sparsity promotion leads to the ASR performance significantly different from SA, e.g., $\mathrm{ASR} = 94.49\%$ vs. $\mathrm{SA} = 11.38\%$ in the top plot of Fig.~\ref{fig:res_trigger}. And the phenomenon is weakened after fine-tuning the Trojan ticket, as indicated by the reduced ASR in Fig.~\ref{fig:res_trigger}. These observations yield two implications. First, the Trojan model exhibits a `fingerprint' in the extreme sparsity regime, where ASR is preserved but SA reduces to the nearly-random performance (because of this extreme  pruning level). Such a fingerprint is called \textit{winning Trojan ticket} termed in Sec.\,\ref{sec: Method} due to its high ASR. Second, this superior Trojan behavior is not well-maintained after the weight finetuning, suggesting that the Trojan effect is mostly encoded by the sparse pattern of the winning Trojan ticket. We also visualize the loss landscape of winning Trojan tickets in Appendix~\ref{sec:more_results}. Last but not the least, the winning Trojan ticket is associated with the peak Trojan score \eqref{eq: S_Troj}, which can thus be leveraged as a powerful tool for Trojan detection. 

\subsubsection{Backdoor properties of winning Trojan ticket}

In Fig.\,\ref{fig:norm}, we next investigate the backdoor properties embedded in the \textit{winning Trojan ticket}, which is identified by the peak Trojan score (see examples in Fig.~\ref{fig:res_trigger}). Our \textbf{key findings} are summarized below. \textbf{({i})} Among dense and various sparse networks, the winning Trojan ticket needs the \textit{minimum  perturbation} to reverse engineering of the Trojan targeted label $t_{\mathrm{Trojan}}$ found by \eqref{eq: RED_trojan}. The performance of our approach outperforms the baseline method, named Neural Cleanse (NC) \cite{wang2019neural}. \textbf{({ii})} The recovered   trigger pattern $(z^*(t_{\mathrm{Trojan}}) \odot \delta^*(t_{\mathrm{Trojan}}) )$ using \eqref{eq: RED_trojan} indeed yields a valid Trojan attack of high ASR. \textbf{(iii)} By leveraging the  winning Trojan   ticket, we can achieve the Trojan trigger recovery for `free'. That is, the high-quality Trojan attack can be recovered using only `noise image inputs' when solving the problem \eqref{eq: RED_trojan}. We highlight that the aforementioned findings (i)-(iii) are consistent across different Trojan attack types, datasets, and model architectures.

In each sub-plot of Fig.\,\ref{fig:norm}, we demonstrate the $\ell_1$ norm of the recovered Trojan trigger $(z^*(t) \odot \delta^*(t) )$ by solving the problem \eqref{eq: RED_trojan} at different specifications of the class label $t$ and the victim model $\phi$. We enumerate all the possible choices of $t$ and examine three types of victim models, given by the winning Trojan ticket (with the peak Trojan score), the originally dense Trojan model (used by NC \cite{wang2019neural}), and the non-Trojan dense model  (that is normally-trained over the benign training dataset). Multiple sub-plots of Fig.\,\ref{fig:norm} correspond to our experiments across different model architectures, different ground-truth Trojan trigger types, and different input images used to solve  problem \eqref{eq: RED_trojan}.
It is clear from Fig.\,\ref{fig:norm} that in all experiments, our identified   Trojan ticket yields the least perturbation norm of the recovered Trojan trigger  at the Trojan targeted label (\textit{i.e.}, $t = t_{\mathrm{Trojan}}$).
The rationale behind the \textit{minimum perturbation criterion}  is that if there exists a backdoor `shortcut'  in the Trojan model (with high ASR), then an input image  only needs the very tiny perturbation   optimized towards $t = t_{\mathrm{Trojan}}$ \cite{wang2019neural}. As a result, one can detect the target label by  just monitoring the perturbation norm. Moreover, we observe that the baseline NC method (associated with the dense Trojan model) \cite{wang2019neural}  lacks stability. For example, it fails to identify the correct target label at the use of the RGB trigger (e.g., Fig.\,\ref{fig:norm} [d]). 
Further, we note that  the non-Trojan model does not follow the minimum perturbation-based detection rule.   

\begin{table}[!ht]
\centering
\caption{Performance of recovered triggers with ResNet-20s on CIFAR-10 across diverse Trojan triggers, including gray-scale, RGB, and clean-label triggers. \cmark/\xmark mean the detected label is matched/unmatched with the true target label. }
\label{tab:asr_triggers}
\vspace{-3mm}
\resizebox{1\linewidth}{!}{
\begin{tabular}{l|cc}
\toprule
Gray-scale Trigger & (Detected, $\ell_1$) & ASR \\ \midrule
Dense baseline~\cite{guo2019tabor} & (``$1$", $196.8$) \cmark & $71.4$\% \\ 
Winning Trojan ticket & (``$1$", $68.0$) \cmark & $\mathbf{91.2}$\% \\ 
\bottomrule
\end{tabular}}
\resizebox{1\linewidth}{!}{
\begin{tabular}{l|cc}
\toprule
RGB Trigger & (Detected, $\ell_1$) & ASR  \\ \midrule
Dense baseline~\cite{guo2019tabor} & (``$1$", $78.7$) \cmark & $48.0$\% \\ 
Winning Trojan ticket & (``$1$", $29.8$) \cmark & $\mathbf{99.6}$\% \\ 
\bottomrule
\end{tabular}}
\resizebox{1\linewidth}{!}{
\begin{tabular}{l|cc}
\toprule
Clean-label Trigger & (Detected, $\ell_1$) & ASR \\ \midrule
Dense baseline~\cite{guo2019tabor} & (``$1$", $48.6$) \cmark & $9.6$\% \\ 
Winning Trojan ticket & (``$1$", $14.0$) \cmark & $\mathbf{99.8}$\% \\ 
\bottomrule
\end{tabular}}
\vspace{-2mm}
\end{table}

\begin{table}[!ht]
\centering
\caption{Performance of recovered triggers with RGB Trojan attack across diverse combinations of network architectures and datasets, i.e., (Vgg-16, CIFAR-10), (ResNet-20s, CIFAR-100), (ResNet-18, R-ImageNet).}
\label{tab:asr_arch_data}
\vspace{-3mm}
\resizebox{1\linewidth}{!}{
\begin{tabular}{l|cc}
\toprule
(VGG-16, CIFAR-10) & (Detected, $\ell_1$) & ASR \\ \midrule
Dense baseline~\cite{guo2019tabor} & (``$1$", $83.3$) \cmark & $33.6$\% \\ 
Winning Trojan ticket & (``$1$", $15.0$) \cmark & $\mathbf{100.0}$\% \\ 
\bottomrule
\end{tabular}}
\resizebox{1\linewidth}{!}{
\begin{tabular}{l|cc}
\toprule
(ResNet-20s, CIFAR-100) & (Detected, $\ell_1$) & ASR \\ \midrule
Dense baseline~\cite{guo2019tabor} & (``$1$", $149.9$) \cmark & $13.8$ \\
Winning Trojan ticket & (``$1$", $132.7$) \cmark & $\mathbf{98.7}$ \\ 
\bottomrule
\end{tabular}}
\resizebox{1\linewidth}{!}{
\begin{tabular}{l|cc}
\toprule
(ResNet-18, R-ImageNet) & (Detected, $\ell_1$) & ASR  \\ \midrule
Dense baseline~\cite{guo2019tabor} & (``$9$", $13.9$) \xmark & $9.8$  \\ 
Winning Trojan ticket & (``$1$", $193.1$) \cmark & $\mathbf{98.7}$ \\ 
\bottomrule
\end{tabular}}
\end{table}
 
In Tab.\,\ref{tab:asr_triggers},  we present the attack performance (ASR) of the recovered Trojan trigger  versus  the different choice of the ground-truth Trojan trigger type (\textit{i.e.}, gray-scale, RGB, and clean-label trigger).  As we can see, even if the baseline NC method (associated with the dense Trojan model) can correctly identify the target label, the quality of the recovered Trojan trigger is poor, justified by its much lower ASR than ours. In particular, when the clean-label attack was used in the Trojan model, our approach (by leveraging the winning Trojan ticket) leads to over $90\%$ ASR improvement. 
In Tab.\,\ref{tab:asr_arch_data}, we present the ASR of the recovered Trojan trigger under different model architectures and datasets. Consistent with Tab.\,\ref{tab:asr_triggers}, the use of the winning Trojan ticket significantly outperforms the baseline approach, not only in ASR but also in the correctness of the detected target label based on the minimum perturbation criterion.

\begin{figure}[t]
    \centering
    \includegraphics[width=0.98\linewidth]{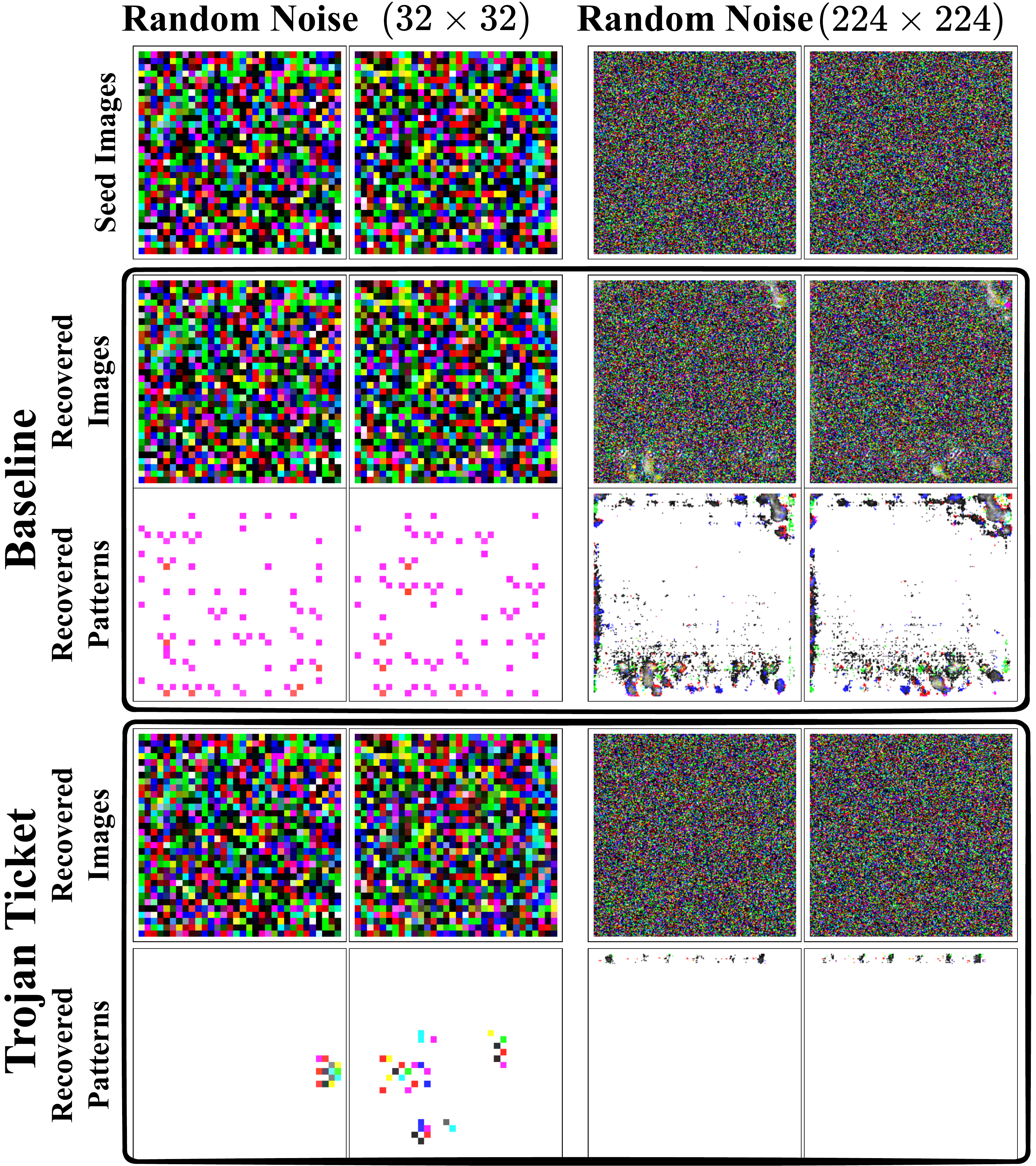}
    \vspace{-2mm}
    \caption{Visualization of recovered Trojan trigger patterns from dense Trojan models and winning Trojan tickets. ResNet-20s on CIFAR-10 and ResNet-18 on ImageNet with RGB triggers are used here. The first row shows the random base images used for solving the problem~\eqref{eq: RED_trojan}, which is a challenging scheme from~\cite{wang2020practical}.}
    \vspace{-2mm}
    \label{fig:recovery}
\end{figure}

In Tab.\,\ref{tab:clean_image}, we examine how the choice of base images in the Trojan recovery problem \eqref{eq: RED_trojan} affects the estimated Trojan quality.  In contrast to the use of $100$ noise images randomly drawn from the standard Gaussian distribution, we also consider the case of using $100$ clean images drawn from the benign data distribution.  As we can see, our approach based on the winning Trojan ticket yields superior Trojan recovery performance to the baseline method in both settings of base images. Most importantly, the quality of our recovered Trojan trigger is  input-agnostic: The $99.6\%$ ASR is achieved using just noise images without having access to any benign images. This is a promising finding of Trojan recovery  `for free' given the zero knowledge about how the Trojan attack is injected into the model training pipeline. The superiority of our approach can also be justified from the visualized Trojan trigger estimates  in Fig.\,\ref{fig:recovery}. Compared to the baseline NC \cite{wang2019neural}, the more clustered and the sparser Trojan trigger is achieved with much higher ASR shown in Tab.\,\ref{tab:clean_image}. Moreover, we remark that compared to \cite{sun2020poisoned} which needs human intervention to craft the sparse trigger estimate, ours provides an automatic way to reverse engineer the valid and the sparse Trojan trigger.

\begin{table}[t]
\centering
\caption{Performance of recovered triggers with random noise images (`free') v.s. benign clean images. The RGB Trojan attack on CIFAR-10 and ResNet-20s are used for the reverse engineering.}
\label{tab:clean_image}
\vspace{-3mm}
\resizebox{1\linewidth}{!}{
\begin{tabular}{l|cc}
\toprule
Noise Images (`Free') & (Detected, $\ell_1$) & ASR \\ \midrule
Dense baseline~\cite{guo2019tabor} & (``$1$", $78.7$) \cmark & $48.0$\% \\ 
Winning Trojan ticket & (``$1$", $29.8$) \cmark & $\mathbf{99.6}$\% \\ 
\bottomrule
\end{tabular}}
\resizebox{1\linewidth}{!}{
\begin{tabular}{l|cc}
\toprule
Clean Images & (Detected, $\ell_1$) & ASR \\ \midrule
Dense baseline~\cite{guo2019tabor} & (``$1$", $174.6$) \cmark & $72.6$\% \\
Winning Trojan ticket & (``$1$", $40.4$) \cmark & $\mathbf{99.8}$\% \\ 
\bottomrule
\end{tabular}}
\end{table}

\paragraph{Ablation study.} In Appendix~\ref{sec:more_results}, we provide more ablations on the sensitivity of our proposal to the sparse network selection, the configurations of Trojan triggers and LTH pruning, and other pruning methods. Meanwhile, visualizations of winning Trojan tickets' sparse connectivities and loss landscape geometry are also presented. Lastly, we further offer extra experiment results on advanced Trojan attackers~\cite{nguyen2021wanet}, more poisoned and un-poisoned datasets.

\section{Conclusion and Discussion}
This paper as pioneering research bridges the lottery ticket hypothesis towards the goal of Trojan trigger detection without any available clean data by a two-step decomposition of first locating a \textit{winning Trojan ticket} with nearly full backdoor and little clean information; then leveraging it to recover the trigger patterns. The effectiveness of our proposals is comprehensively validated across trigger types, network architecture, and datasets.


As the existence of backdoor attacks has aroused increasing public concern on the safe adoption of third-party models, this method provides model suppliers (like the Caffe Model Zoo) with an effective way to inspect the to-be-released models while not requiring any other clean dataset. 
Nevertheless, we admit pruning indeed slows down the pipeline and in our future work, we seek to provide a more computationally efficient method, that can scale up to larger and deeper models. This work is designed to defend malicious attackers, but it might also be abused, which can be constrained by issuing strict licenses.

\vspace{-1mm}
\section*{Acknowledgement}
\vspace{-1mm}
The work of Y. Zhang and S. Liu was supported by the MIT-IBM Watson AI Lab, IBM Research. Z. Wang was in part supported by the NSF grant \#2133861.

\clearpage

{\small
\bibliographystyle{ieee_fullname}
\bibliography{BP}

\begin{thebibliography}{10}\itemsep=-1pt

\bibitem{biggio2013evasion}
Battista Biggio, Igino Corona, Davide Maiorca, Blaine Nelson, Nedim
  {\v{S}}rndi{\'c}, Pavel Laskov, Giorgio Giacinto, and Fabio Roli.
\newblock Evasion attacks against machine learning at test time.
\newblock In {\em Joint European conference on machine learning and knowledge
  discovery in databases}, pages 387--402. Springer, 2013.

\bibitem{boyd2011distributed}
Stephen Boyd, Neal Parikh, and Eric Chu.
\newblock {\em Distributed optimization and statistical learning via the
  alternating direction method of multipliers}.
\newblock Now Publishers Inc, 2011.

\bibitem{carlini2017towards}
Nicholas Carlini and David Wagner.
\newblock Towards evaluating the robustness of neural networks.
\newblock In {\em 2017 ieee symposium on security and privacy (sp)}, pages
  39--57. IEEE, 2017.

\bibitem{chen2018detecting}
Bryant Chen, Wilka Carvalho, Nathalie Baracaldo, Heiko Ludwig, Benjamin
  Edwards, Taesung Lee, Ian Molloy, and Biplav Srivastava.
\newblock Detecting backdoor attacks on deep neural networks by activation
  clustering.
\newblock {\em arXiv preprint arXiv:1811.03728}, 2018.

\bibitem{chen2019deepinspect}
Huili Chen, Cheng Fu, Jishen Zhao, and Farinaz Koushanfar.
\newblock Deepinspect: A black-box trojan detection and mitigation framework
  for deep neural networks.
\newblock In {\em IJCAI}, pages 4658--4664, 2019.

\bibitem{chen2017deeplab}
Liang-Chieh Chen, George Papandreou, Iasonas Kokkinos, Kevin Murphy, and Alan~L
  Yuille.
\newblock Deeplab: Semantic image segmentation with deep convolutional nets,
  atrous convolution, and fully connected crfs.
\newblock {\em IEEE transactions on pattern analysis and machine intelligence},
  40(4):834--848, 2017.

\bibitem{chen2021ultra}
Tianlong Chen, Yu Cheng, Zhe Gan, Jingjing Liu, and Zhangyang Wang.
\newblock Ultra-data-efficient gan training: Drawing a lottery ticket first,
  then training it toughly.
\newblock {\em arXiv preprint arXiv:2103.00397}, 2021.

\bibitem{chen2020lottery2}
Tianlong Chen, Jonathan Frankle, Shiyu Chang, Sijia Liu, Yang Zhang, Michael
  Carbin, and Zhangyang Wang.
\newblock The lottery tickets hypothesis for supervised and self-supervised
  pre-training in computer vision models.
\newblock {\em arXiv preprint arXiv:2012.06908}, 2020.

\bibitem{chen2020lottery}
Tianlong Chen, Jonathan Frankle, Shiyu Chang, Sijia Liu, Yang Zhang, Zhangyang
  Wang, and Michael Carbin.
\newblock The lottery ticket hypothesis for pre-trained bert networks, 2020.

\bibitem{chen2021unified}
Tianlong Chen, Yongduo Sui, Xuxi Chen, Aston Zhang, and Zhangyang Wang.
\newblock A unified lottery ticket hypothesis for graph neural networks.
\newblock {\em arXiv preprint arXiv:2102.06790}, 2021.

\bibitem{chen2017targeted}
Xinyun Chen, Chang Liu, Bo Li, Kimberly Lu, and Dawn Song.
\newblock Targeted backdoor attacks on deep learning systems using data
  poisoning, 2017.

\bibitem{chen2021gans}
Xuxi Chen, Zhenyu Zhang, Yongduo Sui, and Tianlong Chen.
\newblock {\{}GAN{\}}s can play lottery tickets too.
\newblock In {\em International Conference on Learning Representations}, 2021.

\bibitem{deng2009imagenet}
Jia Deng, Wei Dong, Richard Socher, Li-Jia Li, Kai Li, and Li Fei-Fei.
\newblock Imagenet: A large-scale hierarchical image database.
\newblock In {\em 2009 IEEE conference on computer vision and pattern
  recognition}, pages 248--255. Ieee, 2009.

\bibitem{deng2012mnist}
Li Deng.
\newblock The mnist database of handwritten digit images for machine learning
  research.
\newblock {\em IEEE Signal Processing Magazine}, 29(6):141--142, 2012.

\bibitem{dhillon2018stochastic}
Guneet~S Dhillon, Kamyar Azizzadenesheli, Zachary~C Lipton, Jeremy Bernstein,
  Jean Kossaifi, Aran Khanna, and Anima Anandkumar.
\newblock Stochastic activation pruning for robust adversarial defense.
\newblock {\em arXiv preprint arXiv:1803.01442}, 2018.

\bibitem{doan2020februus}
Bao~Gia Doan, Ehsan Abbasnejad, and Damith~C Ranasinghe.
\newblock Februus: Input purification defense against trojan attacks on deep
  neural network systems.
\newblock In {\em Annual Computer Security Applications Conference}, pages
  897--912, 2020.

\bibitem{draxler2018essentially}
Felix Draxler, Kambis Veschgini, Manfred Salmhofer, and Fred Hamprecht.
\newblock Essentially no barriers in neural network energy landscape.
\newblock In {\em International conference on machine learning}, pages
  1309--1318. PMLR, 2018.

\bibitem{frankle2018lottery}
Jonathan Frankle and Michael Carbin.
\newblock The lottery ticket hypothesis: Finding sparse, trainable neural
  networks.
\newblock {\em arXiv preprint arXiv:1803.03635}, 2018.

\bibitem{frankle2020linear}
Jonathan Frankle, Gintare~Karolina Dziugaite, Daniel Roy, and Michael Carbin.
\newblock Linear mode connectivity and the lottery ticket hypothesis.
\newblock In {\em International Conference on Machine Learning}, pages
  3259--3269. PMLR, 2020.

\bibitem{frankle2020pruning}
Jonathan Frankle, Gintare~Karolina Dziugaite, Daniel~M Roy, and Michael Carbin.
\newblock Pruning neural networks at initialization: Why are we missing the
  mark?
\newblock {\em arXiv preprint arXiv:2009.08576}, 2020.

\bibitem{gale2019state}
Trevor Gale, Erich Elsen, and Sara Hooker.
\newblock The state of sparsity in deep neural networks.
\newblock {\em arXiv}, abs/1902.09574, 2019.

\bibitem{gan2021playing}
Zhe Gan, Yen-Chun Chen, Linjie Li, Tianlong Chen, Yu Cheng, Shuohang Wang, and
  Jingjing Liu.
\newblock Playing lottery tickets with vision and language.
\newblock {\em arXiv preprint arXiv:2104.11832}, 2021.

\bibitem{gao2017deepcloak}
Ji Gao, Beilun Wang, Zeming Lin, Weilin Xu, and Yanjun Qi.
\newblock Deepcloak: Masking deep neural network models for robustness against
  adversarial samples.
\newblock {\em arXiv preprint arXiv:1702.06763}, 2017.

\bibitem{gao2020strip}
Yansong Gao, Chang Xu, Derui Wang, Shiping Chen, Damith~C. Ranasinghe, and
  Surya Nepal.
\newblock Strip: A defence against trojan attacks on deep neural networks,
  2020.

\bibitem{garipov2018loss}
Timur Garipov, Pavel Izmailov, Dmitrii Podoprikhin, Dmitry Vetrov, and
  Andrew~Gordon Wilson.
\newblock Loss surfaces, mode connectivity, and fast ensembling of dnns.
\newblock In {\em Proceedings of the 32nd International Conference on Neural
  Information Processing Systems}, pages 8803--8812, 2018.

\bibitem{goldblum2020data}
Micah Goldblum, Dimitris Tsipras, Chulin Xie, Xinyun Chen, Avi Schwarzschild,
  Dawn Song, Aleksander Madry, Bo Li, and Tom Goldstein.
\newblock Data security for machine learning: Data poisoning, backdoor attacks,
  and defenses.
\newblock {\em arXiv preprint arXiv:2012.10544}, 2020.

\bibitem{goldblum2020dataset}
Micah Goldblum, Dimitris Tsipras, Chulin Xie, Xinyun Chen, Avi Schwarzschild,
  Dawn Song, Aleksander Madry, Bo Li, and Tom Goldstein.
\newblock Dataset security for machine learning: Data poisoning, backdoor
  attacks, and defenses.
\newblock {\em arXiv preprint arXiv:2012.10544}, 2020.

\bibitem{goodfellow2014generative}
Ian Goodfellow, Jean Pouget-Abadie, Mehdi Mirza, Bing Xu, David Warde-Farley,
  Sherjil Ozair, Aaron Courville, and Yoshua Bengio.
\newblock Generative adversarial nets.
\newblock {\em Advances in neural information processing systems}, 27, 2014.

\bibitem{gu2017badnets}
Tianyu Gu, Brendan Dolan-Gavitt, and Siddharth Garg.
\newblock Badnets: Identifying vulnerabilities in the machine learning model
  supply chain.
\newblock {\em arXiv preprint arXiv:1708.06733}, 2017.

\bibitem{gu2019badnets}
Tianyu Gu, Brendan Dolan-Gavitt, and Siddharth Garg.
\newblock Badnets: Identifying vulnerabilities in the machine learning model
  supply chain, 2019.

\bibitem{gui2019model}
Shupeng Gui, Haotao~N Wang, Haichuan Yang, Chen Yu, Zhangyang Wang, and Ji Liu.
\newblock Model compression with adversarial robustness: A unified optimization
  framework.
\newblock {\em Advances in Neural Information Processing Systems},
  32:1285--1296, 2019.

\bibitem{guo2019tabor}
Wenbo Guo, Lun Wang, Xinyu Xing, Min Du, and Dawn Song.
\newblock Tabor: A highly accurate approach to inspecting and restoring trojan
  backdoors in ai systems.
\newblock {\em arXiv preprint arXiv:1908.01763}, 2019.

\bibitem{guo2021gdp}
Yi Guo, Huan Yuan, Jianchao Tan, Zhangyang Wang, Sen Yang, and Ji Liu.
\newblock Gdp: Stabilized neural network pruning via gates with differentiable
  polarization.
\newblock In {\em Proceedings of the IEEE/CVF International Conference on
  Computer Vision}, pages 5239--5250, 2021.

\bibitem{guo2018sparse}
Yiwen Guo, Chao Zhang, Changshui Zhang, and Yurong Chen.
\newblock Sparse dnns with improved adversarial robustness.
\newblock {\em arXiv preprint arXiv:1810.09619}, 2018.

\bibitem{han2015deep}
Song Han, Huizi Mao, and William~J Dally.
\newblock Deep compression: Compressing deep neural networks with pruning,
  trained quantization and huffman coding.
\newblock {\em arXiv preprint arXiv:1510.00149}, 2015.

\bibitem{han2015learning}
Song Han, Jeff Pool, John Tran, and William~J Dally.
\newblock Learning both weights and connections for efficient neural network.
\newblock In {\em NIPS}, 2015.

\bibitem{hassibi1993second}
Babak Hassibi and David~G Stork.
\newblock {\em Second order derivatives for network pruning: Optimal brain
  surgeon}.
\newblock Morgan Kaufmann, 1993.

\bibitem{he2016deep}
Kaiming He, Xiangyu Zhang, Shaoqing Ren, and Jian Sun.
\newblock Deep residual learning for image recognition.
\newblock In {\em Proceedings of the IEEE conference on computer vision and
  pattern recognition}, pages 770--778, 2016.

\bibitem{he2017channel}
Yihui He, Xiangyu Zhang, and Jian Sun.
\newblock Channel pruning for accelerating very deep neural networks.
\newblock In {\em Proceedings of the IEEE International Conference on Computer
  Vision}, 2017.

\bibitem{hong2021handcrafted}
Sanghyun Hong, Nicholas Carlini, and Alexey Kurakin.
\newblock Handcrafted backdoors in deep neural networks.
\newblock {\em arXiv preprint arXiv:2106.04690}, 2021.

\bibitem{Houben-IJCNN-2013}
Sebastian Houben, Johannes Stallkamp, Jan Salmen, Marc Schlipsing, and
  Christian Igel.
\newblock Detection of traffic signs in real-world images: The {G}erman
  {T}raffic {S}ign {D}etection {B}enchmark.
\newblock In {\em International Joint Conference on Neural Networks}, 2013.

\bibitem{huang2017densely}
Gao Huang, Zhuang Liu, Laurens Van Der~Maaten, and Kilian~Q Weinberger.
\newblock Densely connected convolutional networks.
\newblock In {\em Proceedings of the IEEE conference on computer vision and
  pattern recognition}, pages 4700--4708, 2017.

\bibitem{jagielski2018manipulating}
Matthew Jagielski, Alina Oprea, Battista Biggio, Chang Liu, Cristina
  Nita-Rotaru, and Bo Li.
\newblock Manipulating machine learning: Poisoning attacks and countermeasures
  for regression learning.
\newblock In {\em 2018 IEEE Symposium on Security and Privacy (SP)}, pages
  19--35. IEEE, 2018.

\bibitem{janowsky1989pruning}
Steven~A Janowsky.
\newblock Pruning versus clipping in neural networks.
\newblock {\em Physical Review A}, 39(12):6600, 1989.

\bibitem{kalibhat2021winning}
Neha~Mukund Kalibhat, Yogesh Balaji, and Soheil Feizi.
\newblock Winning lottery tickets in deep generative models, 2021.

\bibitem{krizhevsky2009learning}
Alex Krizhevsky, Geoffrey Hinton, et~al.
\newblock Learning multiple layers of features from tiny images.
\newblock 2009.

\bibitem{NIPS2012_c399862d}
Alex Krizhevsky, Ilya Sutskever, and Geoffrey~E Hinton.
\newblock Imagenet classification with deep convolutional neural networks.
\newblock In F. Pereira, C.~J.~C. Burges, L. Bottou, and K.~Q. Weinberger,
  editors, {\em Advances in Neural Information Processing Systems}, volume~25.
  Curran Associates, Inc., 2012.

\bibitem{kurakin2016adversarial}
Alexey Kurakin, Ian Goodfellow, and Samy Bengio.
\newblock Adversarial machine learning at scale.
\newblock {\em arXiv preprint arXiv:1611.01236}, 2016.

\bibitem{lecun1990optimal}
Yann LeCun, John~S Denker, and Sara~A Solla.
\newblock Optimal brain damage.
\newblock In {\em Advances in neural information processing systems}, pages
  598--605, 1990.

\bibitem{lee2018snip}
Namhoon Lee, Thalaiyasingam Ajanthan, and Philip~HS Torr.
\newblock Snip: Single-shot network pruning based on connection sensitivity.
\newblock {\em arXiv preprint arXiv:1810.02340}, 2018.

\bibitem{li2021neural}
Yige Li, Xixiang Lyu, Nodens Koren, Lingjuan Lyu, Bo Li, and Xingjun Ma.
\newblock Neural attention distillation: Erasing backdoor triggers from deep
  neural networks.
\newblock {\em arXiv preprint arXiv:2101.05930}, 2021.

\bibitem{li2020rethinking}
Yiming Li, Tongqing Zhai, Baoyuan Wu, Yong Jiang, Zhifeng Li, and Shutao Xia.
\newblock Rethinking the trigger of backdoor attack.
\newblock {\em arXiv preprint arXiv:2004.04692}, 2020.

\bibitem{liu2018fine}
Kang Liu, Brendan Dolan-Gavitt, and Siddharth Garg.
\newblock Fine-pruning: Defending against backdooring attacks on deep neural
  networks.
\newblock In {\em International Symposium on Research in Attacks, Intrusions,
  and Defenses}, pages 273--294. Springer, 2018.

\bibitem{liu2017trojaning}
Yingqi Liu, Shiqing Ma, Yousra Aafer, Wen-Chuan Lee, Juan Zhai, Weihang Wang,
  and Xiangyu Zhang.
\newblock Trojaning attack on neural networks.
\newblock 2017.

\bibitem{liu2020reflection}
Yunfei Liu, Xingjun Ma, James Bailey, and Feng Lu.
\newblock Reflection backdoor: A natural backdoor attack on deep neural
  networks.
\newblock In {\em European Conference on Computer Vision}, pages 182--199.
  Springer, 2020.

\bibitem{liu2017learning}
Zhuang Liu, Jianguo Li, Zhiqiang Shen, Gao Huang, Shoumeng Yan, and Changshui
  Zhang.
\newblock Learning efficient convolutional networks through network slimming.
\newblock In {\em Proceedings of the IEEE International Conference on Computer
  Vision}, pages 2736--2744, 2017.

\bibitem{ma2021good}
Haoyu Ma, Tianlong Chen, Ting-Kuei Hu, Chenyu You, Xiaohui Xie, and Zhangyang
  Wang.
\newblock Good students play big lottery better.
\newblock {\em arXiv preprint arXiv:2101.03255}, 2021.

\bibitem{madry2017towards}
Aleksander Madry, Aleksandar Makelov, Ludwig Schmidt, Dimitris Tsipras, and
  Adrian Vladu.
\newblock Towards deep learning models resistant to adversarial attacks.
\newblock {\em arXiv preprint arXiv:1706.06083}, 2017.

\bibitem{molchanov2019importance}
Pavlo Molchanov, Arun Mallya, Stephen Tyree, Iuri Frosio, and Jan Kautz.
\newblock Importance estimation for neural network pruning.
\newblock In {\em Proceedings of the IEEE Conference on Computer Vision and
  Pattern Recognition}, pages 11264--11272, 2019.

\bibitem{molchanov2016pruning}
Pavlo Molchanov, Stephen Tyree, Tero Karras, Timo Aila, and Jan Kautz.
\newblock Pruning convolutional neural networks for resource efficient
  inference.
\newblock {\em arXiv preprint arXiv:1611.06440}, 2016.

\bibitem{mozer1989skeletonization}
Michael~C Mozer and Paul Smolensky.
\newblock Skeletonization: A technique for trimming the fat from a network via
  relevance assessment.
\newblock In {\em Advances in neural information processing systems}, pages
  107--115, 1989.

\bibitem{nguyen2021wanet}
Anh Nguyen and Anh Tran.
\newblock Wanet--imperceptible warping-based backdoor attack.
\newblock {\em arXiv preprint arXiv:2102.10369}, 2021.

\bibitem{ouyang2013stochastic}
Hua Ouyang, Niao He, Long Tran, and Alexander Gray.
\newblock Stochastic alternating direction method of multipliers.
\newblock In {\em International Conference on Machine Learning}, pages 80--88.
  PMLR, 2013.

\bibitem{quiring2020backdooring}
Erwin Quiring and Konrad Rieck.
\newblock Backdooring and poisoning neural networks with image-scaling attacks.
\newblock In {\em 2020 IEEE Security and Privacy Workshops (SPW)}, pages
  41--47. IEEE, 2020.

\bibitem{ren2018admmnn}
Ao Ren, Tianyun Zhang, Shaokai Ye, Jiayu Li, Wenyao Xu, Xuehai Qian, Xue Lin,
  and Yanzhi Wang.
\newblock Admm-nn: An algorithm-hardware co-design framework of dnns using
  alternating direction method of multipliers, 2018.

\bibitem{ren2016faster}
Shaoqing Ren, Kaiming He, Ross Girshick, and Jian Sun.
\newblock Faster r-cnn: Towards real-time object detection with region proposal
  networks.
\newblock {\em IEEE transactions on pattern analysis and machine intelligence},
  39(6):1137--1149, 2016.

\bibitem{Renda2020Comparing}
Alex Renda, Jonathan Frankle, and Michael Carbin.
\newblock Comparing rewinding and fine-tuning in neural network pruning.
\newblock In {\em 8th International Conference on Learning Representations},
  2020.

\bibitem{saha2020hidden}
Aniruddha Saha, Akshayvarun Subramanya, and Hamed Pirsiavash.
\newblock Hidden trigger backdoor attacks.
\newblock In {\em Proceedings of the AAAI Conference on Artificial
  Intelligence}, volume~34, pages 11957--11965, 2020.

\bibitem{schwarzschild2021just}
Avi Schwarzschild, Micah Goldblum, Arjun Gupta, John~P Dickerson, and Tom
  Goldstein.
\newblock Just how toxic is data poisoning? a unified benchmark for backdoor
  and data poisoning attacks.
\newblock In {\em International Conference on Machine Learning}, pages
  9389--9398. PMLR, 2021.

\bibitem{sehwag2019towards}
Vikash Sehwag, Shiqi Wang, Prateek Mittal, and Suman Jana.
\newblock Towards compact and robust deep neural networks.
\newblock {\em arXiv preprint arXiv:1906.06110}, 2019.

\bibitem{shafahi2018poison}
Ali Shafahi, W~Ronny Huang, Mahyar Najibi, Octavian Suciu, Christoph Studer,
  Tudor Dumitras, and Tom Goldstein.
\newblock Poison frogs! targeted clean-label poisoning attacks on neural
  networks.
\newblock {\em arXiv preprint arXiv:1804.00792}, 2018.

\bibitem{shen2019learning}
Yanyao Shen and Sujay Sanghavi.
\newblock Learning with bad training data via iterative trimmed loss
  minimization.
\newblock In {\em International Conference on Machine Learning}, pages
  5739--5748. PMLR, 2019.

\bibitem{simonyan2014very}
Karen Simonyan and Andrew Zisserman.
\newblock Very deep convolutional networks for large-scale image recognition.
\newblock {\em arXiv preprint arXiv:1409.1556}, 2014.

\bibitem{sun2020poisoned}
Mingjie Sun, Siddhant Agarwal, and J~Zico Kolter.
\newblock Poisoned classifiers are not only backdoored, they are fundamentally
  broken.
\newblock {\em arXiv preprint arXiv:2010.09080}, 2020.

\bibitem{tran2018spectral}
Brandon Tran, Jerry Li, and Aleksander Madry.
\newblock Spectral signatures in backdoor attacks.
\newblock {\em arXiv preprint arXiv:1811.00636}, 2018.

\bibitem{tsipras2018robustness}
Dimitris Tsipras, Shibani Santurkar, Logan Engstrom, Alexander Turner, and
  Aleksander Madry.
\newblock Robustness may be at odds with accuracy.
\newblock {\em arXiv preprint arXiv:1805.12152}, 2018.

\bibitem{udeshi2019model}
Sakshi Udeshi, Shanshan Peng, Gerald Woo, Lionell Loh, Louth Rawshan, and
  Sudipta Chattopadhyay.
\newblock Model agnostic defence against backdoor attacks in machine learning.
\newblock {\em arXiv preprint arXiv:1908.02203}, 2019.

\bibitem{villarreal2020confoc}
Miguel Villarreal-Vasquez and Bharat Bhargava.
\newblock Confoc: Content-focus protection against trojan attacks on neural
  networks.
\newblock {\em arXiv preprint arXiv:2007.00711}, 2020.

\bibitem{wang2019neural}
Bolun Wang, Yuanshun Yao, Shawn Shan, Huiying Li, Bimal Viswanath, Haitao
  Zheng, and Ben~Y Zhao.
\newblock Neural cleanse: Identifying and mitigating backdoor attacks in neural
  networks.
\newblock In {\em 2019 IEEE Symposium on Security and Privacy (SP)}, pages
  707--723. IEEE, 2019.

\bibitem{wang2020practical}
Ren Wang, Gaoyuan Zhang, Sijia Liu, Pin-Yu Chen, Jinjun Xiong, and Meng Wang.
\newblock Practical detection of trojan neural networks: Data-limited and
  data-free cases.
\newblock In {\em Computer Vision--ECCV 2020: 16th European Conference,
  Glasgow, UK, August 23--28, 2020, Proceedings, Part XXIII 16}, pages
  222--238. Springer, 2020.

\bibitem{wang2018defending}
Siyue Wang, Xiao Wang, Shaokai Ye, Pu Zhao, and Xue Lin.
\newblock Defending dnn adversarial attacks with pruning and logits
  augmentation.
\newblock In {\em 2018 IEEE Global Conference on Signal and Information
  Processing (GlobalSIP)}, pages 1144--1148. IEEE, 2018.

\bibitem{5995566}
Lior Wolf, Tal Hassner, and Itay Maoz.
\newblock Face recognition in unconstrained videos with matched background
  similarity.
\newblock In {\em CVPR 2011}, pages 529--534, 2011.

\bibitem{wong2020fast}
Eric Wong, Leslie Rice, and J~Zico Kolter.
\newblock Fast is better than free: Revisiting adversarial training.
\newblock {\em arXiv preprint arXiv:2001.03994}, 2020.

\bibitem{wu2021adversarial}
Dongxian Wu and Yisen Wang.
\newblock Adversarial neuron pruning purifies backdoored deep models.
\newblock {\em arXiv preprint arXiv:2110.14430}, 2021.

\bibitem{xiang2020detection}
Zhen Xiang, David~J Miller, and George Kesidis.
\newblock Detection of backdoors in trained classifiers without access to the
  training set.
\newblock {\em IEEE Transactions on Neural Networks and Learning Systems},
  2020.

\bibitem{xu2021defending}
Kaidi Xu, Sijia Liu, Pin-Yu Chen, Pu Zhao, and Xue Lin.
\newblock Defending against backdoor attack on deep neural networks, 2021.

\bibitem{yao2020pyhessian}
Zhewei Yao, Amir Gholami, Kurt Keutzer, and Michael~W Mahoney.
\newblock Pyhessian: Neural networks through the lens of the hessian.
\newblock In {\em 2020 IEEE International Conference on Big Data (Big Data)},
  pages 581--590. IEEE, 2020.

\bibitem{ye2019adversarial}
Shaokai Ye, Kaidi Xu, Sijia Liu, Hao Cheng, Jan-Henrik Lambrechts, Huan Zhang,
  Aojun Zhou, Kaisheng Ma, Yanzhi Wang, and Xue Lin.
\newblock Adversarial robustness vs. model compression, or both?
\newblock In {\em Proceedings of the IEEE/CVF International Conference on
  Computer Vision}, pages 111--120, 2019.

\bibitem{yin2021backdoor}
Zeyuan Yin, Ye Yuan, Panfeng Guo, and Pan Zhou.
\newblock Backdoor attacks on federated learning with lottery ticket
  hypothesis.
\newblock {\em arXiv preprint arXiv:2109.10512}, 2021.

\bibitem{yu2019playing}
Haonan Yu, Sergey Edunov, Yuandong Tian, and Ari~S. Morcos.
\newblock Playing the lottery with rewards and multiple languages: lottery
  tickets in rl and nlp.
\newblock In {\em 8th International Conference on Learning Representations},
  2020.

\bibitem{zhang2019theoretically}
Hongyang Zhang, Yaodong Yu, Jiantao Jiao, Eric Xing, Laurent El~Ghaoui, and
  Michael Jordan.
\newblock Theoretically principled trade-off between robustness and accuracy.
\newblock In {\em International Conference on Machine Learning}, pages
  7472--7482. PMLR, 2019.

\bibitem{pmlr-v139-zhang21c}
Zhenyu Zhang, Xuxi Chen, Tianlong Chen, and Zhangyang Wang.
\newblock Efficient lottery ticket finding: Less data is more.
\newblock In Marina Meila and Tong Zhang, editors, {\em Proceedings of the 38th
  International Conference on Machine Learning}, volume 139 of {\em Proceedings
  of Machine Learning Research}, pages 12380--12390. PMLR, 18--24 Jul 2021.

\bibitem{zhao2020bridging}
Pu Zhao, Pin-Yu Chen, Payel Das, Karthikeyan~Natesan Ramamurthy, and Xue Lin.
\newblock Bridging mode connectivity in loss landscapes and adversarial
  robustness.
\newblock {\em arXiv preprint arXiv:2005.00060}, 2020.

\bibitem{zhao2020clean}
Shihao Zhao, Xingjun Ma, Xiang Zheng, James Bailey, Jingjing Chen, and Yu-Gang
  Jiang.
\newblock Clean-label backdoor attacks on video recognition models.
\newblock In {\em Proceedings of the IEEE/CVF Conference on Computer Vision and
  Pattern Recognition}, pages 14443--14452, 2020.

\bibitem{zhou2016less}
Hao Zhou, Jose~M Alvarez, and Fatih Porikli.
\newblock Less is more: Towards compact cnns.
\newblock In {\em European Conference on Computer Vision}, pages 662--677.
  Springer, 2016.

\bibitem{zhu2019transferable}
Chen Zhu, W~Ronny Huang, Hengduo Li, Gavin Taylor, Christoph Studer, and Tom
  Goldstein.
\newblock Transferable clean-label poisoning attacks on deep neural nets.
\newblock In {\em International Conference on Machine Learning}, pages
  7614--7623. PMLR, 2019.

\end{thebibliography}
}

\clearpage

\appendix

\renewcommand{\thepage}{A\arabic{page}}  
\renewcommand{\thesection}{A\arabic{section}}   
\renewcommand{\thetable}{A\arabic{table}}   
\renewcommand{\thefigure}{A\arabic{figure}}

\section{More Implementation Details} \label{sec:more_implementation}

\paragraph{More details of Trojan attacks.}

\noindent ($1$) Attack configuration for BadNets. We follow the attack methodology proposed in~\cite{gu2017badnets} to inject a backdoor during training. It attaches a trigger with a fixed size ($5 \times 5$) and location (upper right corner) to benign images and injects them into the training set. Specifically, backdoored models are trained on a poisoned dataset with a poison ratio of $1\%$ and the target label is set to $0$ throughout the experiment.

\noindent($2$) Attack configuration for Clean Label Backdoor Attack\cite{zhao2020clean}. This method hinders the model from learning the true salient characteristics of the input through perturbations, often adversarial examples or data generated from GAN. Thus, the learned representations of the images with a target label are distorted towards another class and the content-label mismatch can be achieved in such a manner. In our experiments, we choose the PGD attack~\cite{madry2017towards} to generate adversarial examples for the target class. For each image, we perform a 10-step PGD attack on a robustly trained surrogate ResNet-20s model with an attack budget $\epsilon = 8/255$ and an attack learning rate of $\alpha = 2/255$. The perturbed images are then further attached with the colorful or black trigger as aforementioned. We perturb all the images in the target class to guarantee a successful attack. 

For a recovered trigger ($\mathbf{m},\mathbf{\Delta}$), we evaluate the $ell_1$ norm of soft mask $\mathbf{m}$, and then binarize this mask so that its $ell_1$ norm equals the ground-truth value ($5\times5$ for CIFAR-10/100 and $64\times64$ for R-ImageNet). Then we stamp $\mathbf{\Delta}$ with the binary mask to the test images and calculate the attack successful rate (ASR). 

\vspace{-2mm}
\paragraph{More details of reverse engineering.} 
We use Neural Cleanse~\cite{wang2019neural} as our backbone to conduct trigger reverse engineering. The detection includes two stages. In the first stage, potential triggers with the possibly least norm towards \emph{each} class are obtained through a gradient-descent-based optimization algorithm. The final synthetic trigger and its target label are then determined through an anomaly detector. In the meantime, early stopping is performed as a trick to speed up the trigger recovery.

For trigger recovering, we default to use $100$ noise images generated by Gaussian distribution $\mathcal{N}(0,1)$. And we also compare the quality of recovered triggers from $10$ and $100$ clean images in Table~\ref{tab:clean_image} as an ablation study.

Each time, we pruned $20$\% of the remaining parameters with the lowest magnitude and then rewind the weight to epoch $3$ before retraining.

\section{More Experiment Results} \label{sec:more_results}
In this section, we not only provide comprehensive ablation studies including \ding{172} the fine-tuning steps for Trojan ticket detection; \ding{173} the configurations of Trojan attacks such as the trigger locations; \ding{174} LTH pruning ratios and comparisons with other pruning methods, but also offer \ding{175} visualizations of winning Trojan ticket’s sparse connectivities and loss landscape geometry; \ding{176} pruning dynamics of models with the clean-label Trojan trigger; \ding{177} extra results of stealthier and global triggers; \ding{178} extra results on more datasets; \ding{179} extra results on un-poisoned datasets; \ding{180} failure case analyses.

In addition, for the performance of recovered triggers, we present extra  results of oracle labels (i.e., the truth target class) together with other two sparse Trojan tickets: ($i$) H-Trojan ticket with high SA and ASR; ($ii$) L-Trojan ticket with low SA (standard testing accuracy) and ASR, as collected in Table~\ref{tab:location},~\ref{tab:clean_image_more},~\ref{tab:asr_triggers_more}, and~\ref{tab:asr_arch_more}.

\begin{figure}[!ht]
    \centering
    \includegraphics[width=1\linewidth]{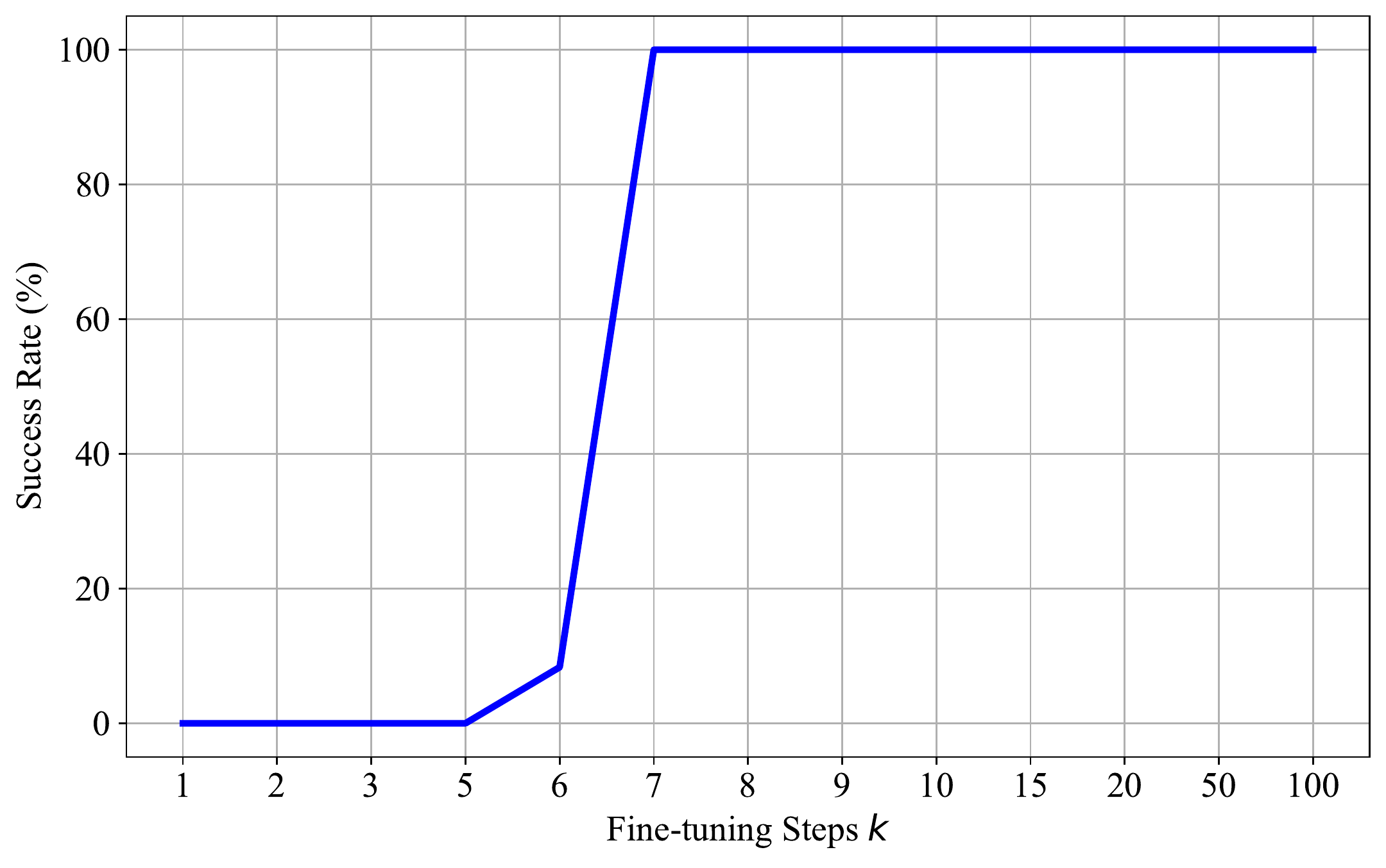}
    \vspace{-8mm}
    \caption{The successful rate over fine-tuning steps of detecting winning Trojan tickets. ResNet-20s on CIFAR-10 with RGB triggers are adopted here.}
    \vspace{-4mm}
    \label{fig:optim_step}
\end{figure}

\vspace{-2mm}
\paragraph{Ablation for the fine-tuning steps $k$.} We explore the effect of the number of fine-tuning steps $k$. The successful rate of detecting winning Trojan tickets is shown in Figure~\ref{fig:optim_step}. It is calculated from \textbf{ten} replicates and each replicate is fine-tuned for $k$ steps. We find that choosing fine-tuning steps $k\ge7$ is potentially enough to accurately identify the winning Trojan tickets.

\begin{figure*}[t]
    \centering
    \includegraphics[width=1\linewidth]{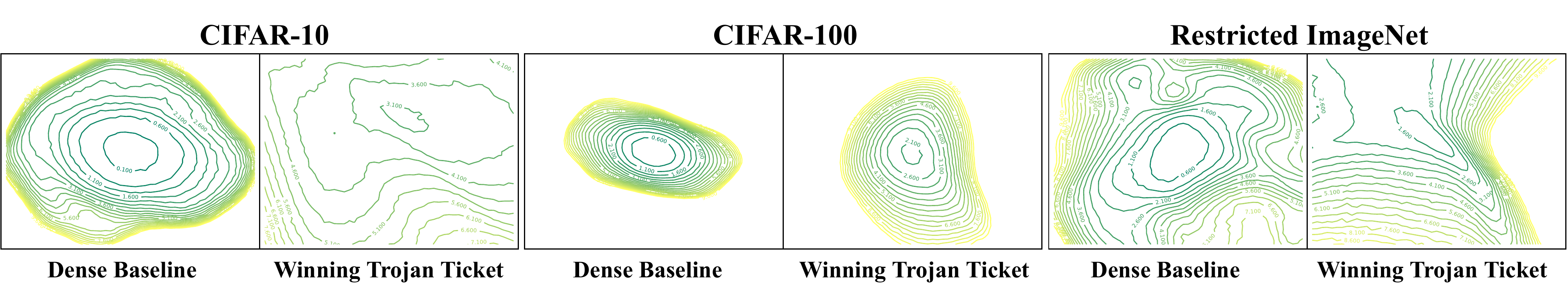}
    \vspace{-6mm}
    \caption{Loss landscape geometry of dense Trojan models and their identified Trojan tickets from CIFAR-10/100, and ImageNet.}
    \vspace{-2mm}
    \label{fig:losssurface}
\end{figure*}

\vspace{-2mm}
\paragraph{Ablation on trigger locations.} We study the different positions for placing Trojan triggers. As shown in Table~\ref{tab:location}, winning Trojan ticket demonstrates a consistent superiority in terms of recovered triggers' ASR.

\begin{table}[!ht]
\centering
\caption{Performance of recovered triggers with ResNet-20s on CIFAR-10. The RGB Trojan attack is applied to different positions, including \textit{bottom left} and \textit{upper right}.}
\label{tab:location}
\vspace{-3mm}
\resizebox{1\linewidth}{!}{
\begin{tabular}{l|cc|cc}
\toprule
\textit{bottom left} & (Detected, $\ell_1$) & ASR & (Oracle, $\ell_1$) & ASR \\ \midrule
Dense baseline~\cite{guo2019tabor} & (``$5$", $121.9$) \xmark & $10.5$\% & (``$1$", $251.6$) & $13.7$\% \\ 
Winning Trojan ticket & (``$1$", $79.3$) \cmark & $\mathbf{86.7}$\% & (``$1$", $79.3$) & $\mathbf{86.7}$\% \\ 
H-Trojan ticket & (``$4$", $104.8$) \xmark & $21.3$\% & (``$1$", $189.1$) & $14.2$\% \\ 
L-Trojan ticket & (``$2$", $158.6$) \xmark & $18.7$\% & (``$1$", $231.3$) & $42.1$\% \\
\bottomrule
\end{tabular}}
\resizebox{1\linewidth}{!}{
\begin{tabular}{l|cc|cc}
\toprule
\textit{upper right} & (Detected, $\ell_1$) & ASR & (Oracle, $\ell_1$) & ASR \\ \midrule
Dense baseline~\cite{guo2019tabor} & (``$1$", $78.7$) \cmark & $48.0$\% & (``$1$", $78.7$) & $48.0$\% \\ 
Winning Trojan ticket & (``$1$", $29.8$) \cmark & $\mathbf{99.6}$\% & (``$1$", $29.8$) & $\mathbf{99.6}$\% \\ 
H-Trojan ticket & (``$7$", $110.9$) \xmark & $8.6$\% & (``$1$", $124.9$) & $18.3$\% \\ 
L-Trojan ticket & (``$2$", $105.0$) \xmark & $58.5$\% & (``$1$", $276.14$) & $17.1$\% \\
\bottomrule
\end{tabular}}
\vspace{-2mm}
\end{table}

\begin{figure}[!ht]
    \centering
    \includegraphics[width=1\linewidth]{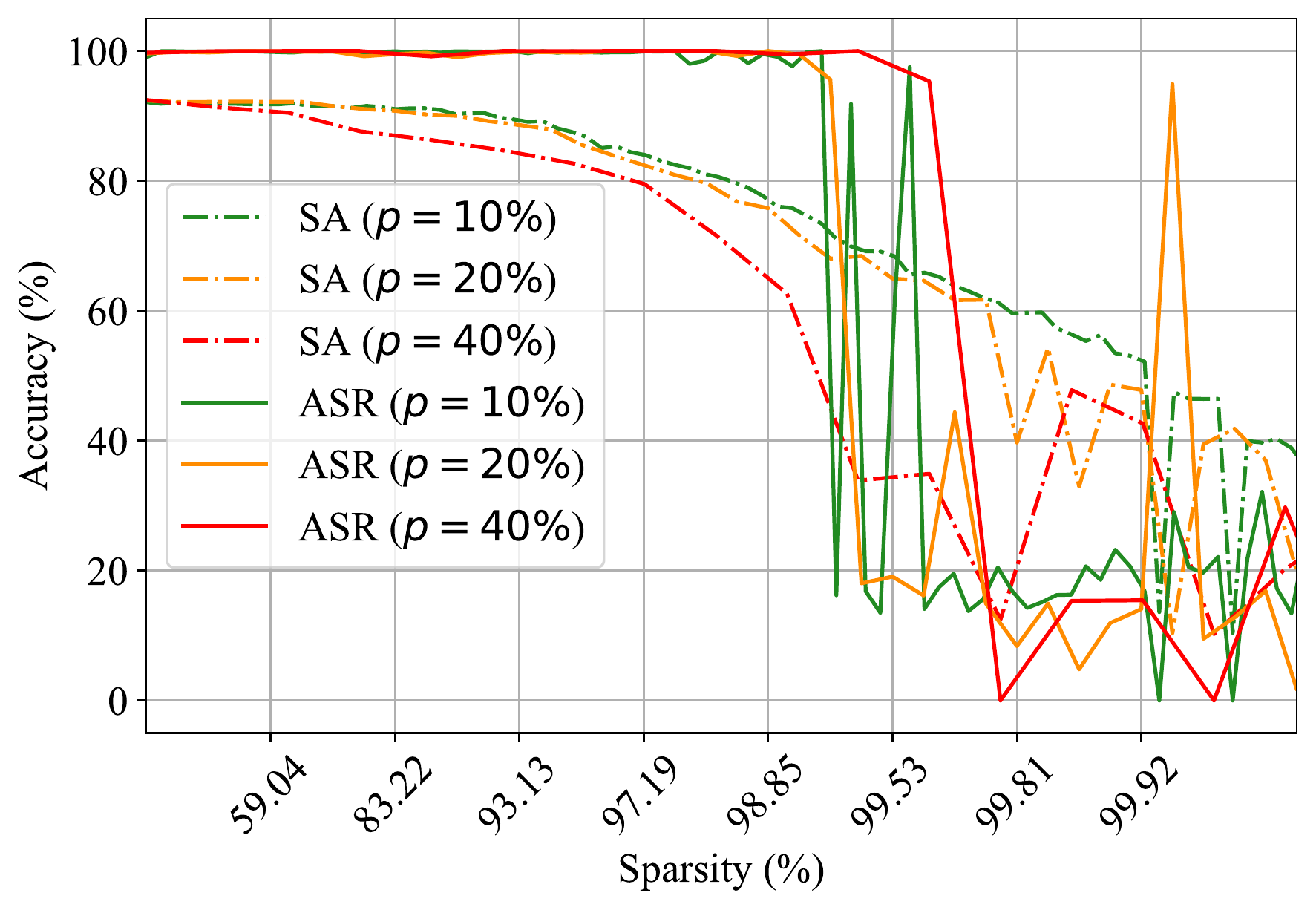}
    \vspace{-8mm}
    \caption{Ablation on the pruning ratio of Trojan ticket findings. The standard testing accuracy (SA \%) and attack successful rate (ASR \%) are reported over network sparsity. ResNet-20s and CIFAR-10 with RGB Trojan tickets are adopted here.}
    \vspace{-4mm}
    \label{fig:aba_ratio}
\end{figure}

\vspace{-2mm}
\paragraph{Ablation on pruning ratios.} We investigate the pruning ratio in LTH pruning~\cite{frankle2018lottery}. Results of pruning ratio $p=10\%$, $20\%$, and $40\%$ are presented in Figure~\ref{fig:aba_ratio}, we observe that $p=10\%$ or $20\%$ are capable of generating the winning Trojan tickets, while $p=40\%$ fails. A possible explanation is that pruning with $p=40\%$ is too aggressive to maintain Trojan information.

\begin{figure}[!ht]
    \centering
    \includegraphics[width=1\linewidth]{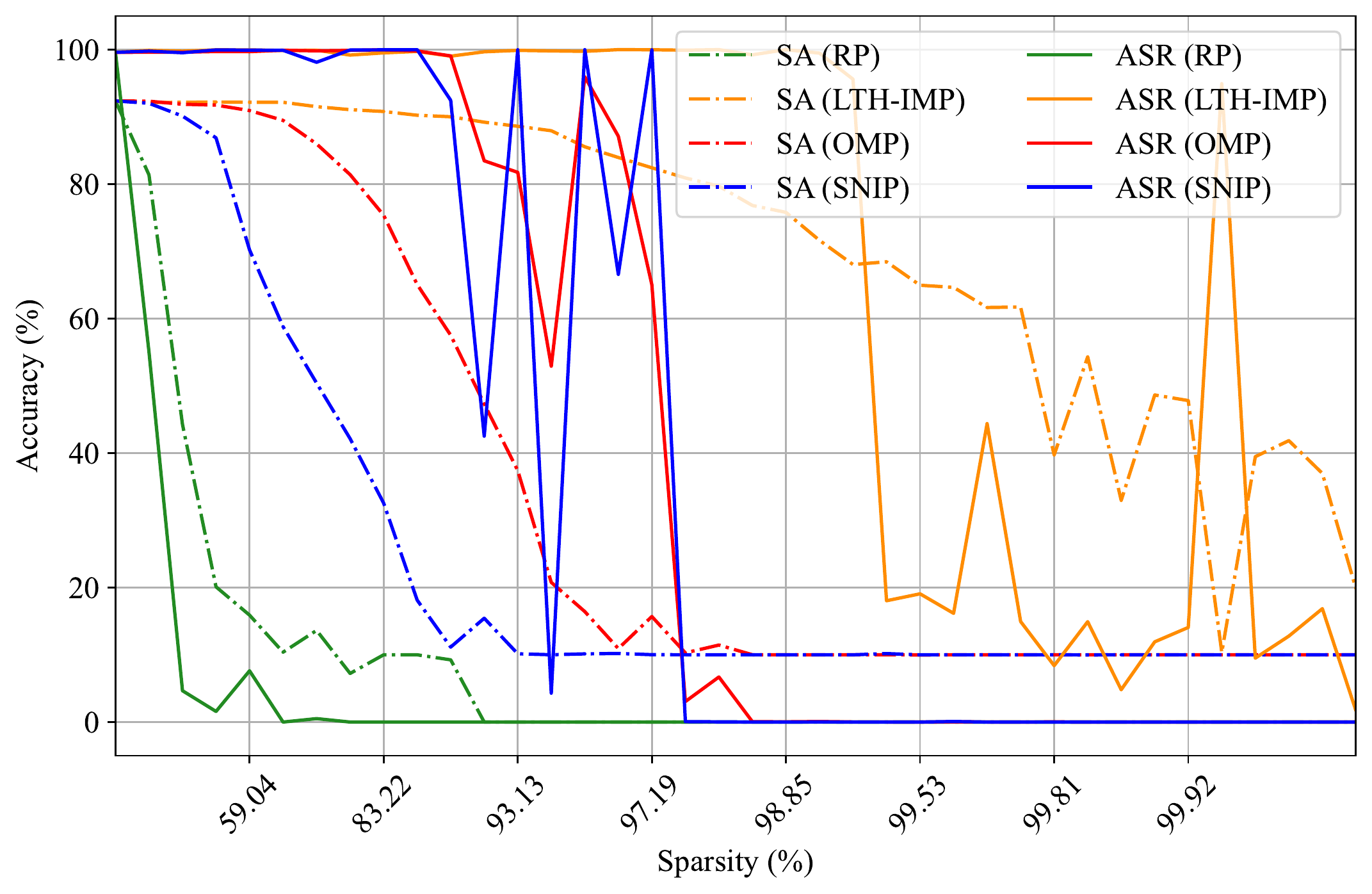}
    \vspace{-6mm}
    \caption{Ablation on the pruning algorithms of Trojan ticket findings, including RP, OMP, GraSP, SNIP and LTH-IMP (ours). The standard testing accuracy (SA \%) and attack successfully rate (ASR \%) are reported over network sparsity.}
    \vspace{-2mm}
    \label{fig:aba_pruning}
\end{figure}

\vspace{-2mm}
\paragraph{Comparison with other pruning methods.} In Figure~\ref{fig:aba_pruning}, we compare LTH pruning~\cite{frankle2018lottery} with other pruning methods like random pruning (RP), one-shot magnitude pruning (OMP), and SNIP~\cite{lee2018snip}. We find that our proposals can be effective across different pruning methods. All of LTH pruning, OMP, and SNIP produce winning Trojan tickets. We also notice that random pruning can not make it, which supports that appropriate sparsity plays a significant role in capturing Trojan information. 

\begin{figure}[!ht]
    \centering
    \includegraphics[width=1\linewidth]{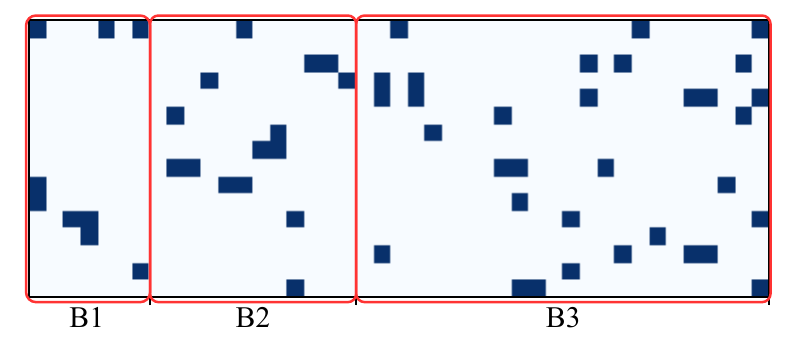}
    \vspace{-8mm}
    \caption{Kernel-wise heatmap visualizations of the winning Trojan ticket with $0.05\%$ sparsity, $11.38\%$ SA, and $94.49\%$ ASR. The bright blocks represent the completely pruned (zero) kernels and the dark  blocks stand for the kernels that have at least one unpruned weight. $\mathrm{B}1\sim3$ donate three residual blocks in the ResNet-20s. CIFAR-10 with RGB triggers is used.}
    \vspace{-1mm}
    \label{fig:mask}
\end{figure}

\vspace{-2mm}
\paragraph{Visualization of sparse masks and loss surfaces.} We visualize the located winning Trojan ticket in Figure~\ref{fig:mask}, and their loss landscape geometries in Figure~\ref{fig:losssurface}. We find that winning Trojan tickets usually have sharp local minima, suggesting a potentially performance gap between before and after fine-tuning which lays the foundation of our proposed detection methods.

\begin{figure}[htb]
    \centering
    \includegraphics[width=1\linewidth]{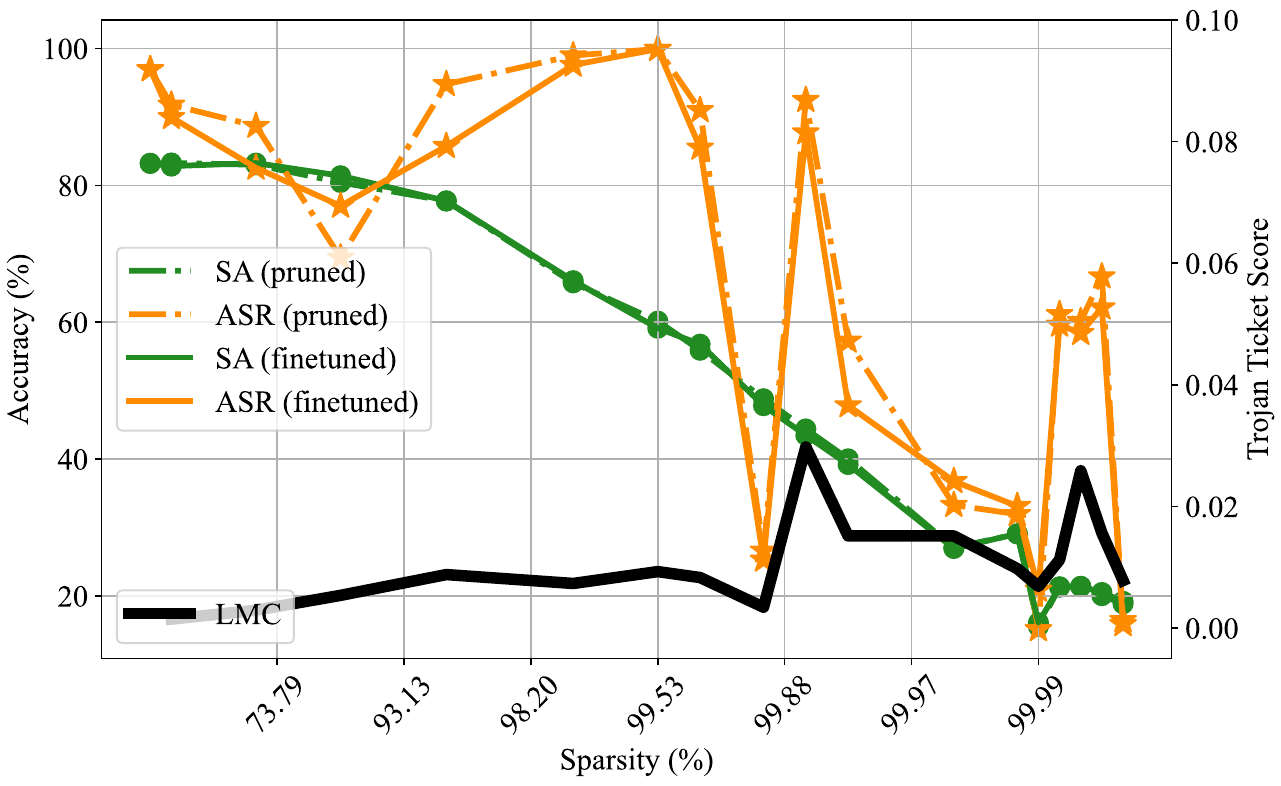}
    \vspace{-8mm}
    \caption{The pruning dynamics and Trojan scores on CIFAR-10 with ResNet-20s using the clean-label Trojan trigger. The peak Trojan score precisely characterizes the winning Trojan ticket.}
    \vspace{-3mm}
    \label{fig:res_trigger_clean}
\end{figure}

\vspace{-2mm}
\paragraph{Pruning dynamic and Trojan scores of the clean-label Trojan trigger.} Figure~\ref{fig:res_trigger_clean} collects the pruning dynamics and Trojan scores on CIFAR-10 dataset with ResNet-20s and the clean-label Trojan trigger, where consistent conclusions can be drown. 

\vspace{-2mm}
\paragraph{Extra results of stealthier and global triggers.} We conduct experiments on another advanced Trojan attack, WaNet~\cite{nguyen2021wanet} which advocates stealthier and global triggers. Results included in below table consistently justified the effectiveness of our approaches. Note that, to enable meaningful comparison, we disable the additive Gaussian noise in~\cite{nguyen2021wanet}; otherwise both dense baseline and winning Trojan tickets are failed in reverse engineering.

\begin{table}[!ht]
\centering
\vspace{-3mm}
\resizebox{1\linewidth}{!}{
\begin{tabular}{c|c|cc}
\toprule
Settings & Noise Images (`Free') & (Detected, $\ell_1$) & ASR \\ \midrule
\multirow{2}{*}{\begin{tabular}[c]{@{}c@{}} (CIFAR-10, ResNet-20s) \\ with \textbf{WaNet} Backdoor \end{tabular}} & Dense baseline & (``$1$", $31.3$) \textcolor{green}{\ding{51}} & $40.1$\% \\ 
 & Winning Trojan ticket & (``$1$", $28.5$) \textcolor{green}{\ding{51}} & $\mathbf{96.71}$\% \\ 
\bottomrule
    \end{tabular}}
\vspace{-4mm}
\end{table}

\vspace{-2mm}
\paragraph{Extra results on more datasets.} Extra experiments are conducted on three more datasets, including MNIST~\cite{deng2012mnist}, GTSRB~\cite{Houben-IJCNN-2013}, and YouTubeFace~\cite{5995566}. Results in the table below reveal similar conclusions as the ones in the main text. For example, as a highly challenging scenario for trigger recovery, the YouTubeFace contains $1283$ classes and the dense baseline method suffers from unsatisfactory results. However, our winning Trojan tickets still succeed in restoring the trigger towards the right target label with a decent ASR.


\begin{table}[!ht]
\centering
\vspace{-3mm}
\resizebox{1\linewidth}{!}{
\begin{tabular}{c|c|cc}
\toprule
Settings & Noise Images (`Free') & (Detected, $\ell_1$) & ASR \\ \midrule
\multirow{2}{*}{\begin{tabular}[c]{@{}c@{}} (\textbf{MNIST}, ResNet-20s) \\ with RGB Triggers \end{tabular}} & Dense baseline & (``$1$", $34.0$) \textcolor{green}{\ding{51}} & $100$\% \\ 
 & Winning Trojan ticket & (``$1$", $36.4$) \textcolor{green}{\ding{51}} & $\mathbf{100}$\% \\ \midrule
\multirow{2}{*}{\begin{tabular}[c]{@{}c@{}} (\textbf{GTSRB}, ResNet-20s) \\ with RGB Triggers \end{tabular}} & Dense baseline & (``$1$", $91.7$) \textcolor{green}{\ding{51}} & $54.02$\% \\ 
 & Winning Trojan ticket & (``$1$", $16.9$) \textcolor{green}{\ding{51}} & $\mathbf{98.89}$\% \\ \midrule
 \multirow{2}{*}{\begin{tabular}[c]{@{}c@{}} (\textbf{YouTube Face}, ResNet-20s) \\ with RGB Triggers \end{tabular}} & Dense baseline & (``$334$", $612.9$) \xmark & $6.23$\% \\ 
 & Winning Trojan ticket & (``$1$", $659.3$) \textcolor{green}{\ding{51}} & $\mathbf{67.03}$\% \\ 
\bottomrule
\end{tabular}}
\vspace{-4mm}
\end{table}

\vspace{-2mm}
\paragraph{Extra results on un-poisoned datasets.} We conducted Trojan detection experiments (in terms of Trojan trigger recovering) on a clean training set. It is shown from the following table that the use of winning Trojan ticket yields similar norms of the recovered triggers across all labels. This will not activate the Trojan detector, and thus, will not flag non-Trojan datasets as the Trojan one. The detection results that we achieved are consistent with~\cite{wang2020practical}.

\begin{table}[!ht]
\centering
\vspace{-3mm}
\resizebox{1\linewidth}{!}{
\begin{tabular}{c|c|cccccccccc}
\toprule
Settings & Label w. $\ell_{\infty}$ & 0 & 1 & 2 & 3 & 4 & 5 & 6 & 7 & 8 & 9 \\ \midrule
\multirow{2}{*}{\begin{tabular}[c]{@{}c@{}} \textbf{Clean} Training Set \\ (CIFAR-10, ResNet-20s) \end{tabular}} & Dense baseline & 254 & 107 & 123 & 126 & 233 & 169 & 187 & 260 & 265 & 207 \\ 
 & Winning Trojan ticket & 436 & 292 & 476 & 379 & 300 & 303 & 301 & 240 & 392 & 259 \\
\bottomrule
\end{tabular}}
\vspace{-4mm}
\end{table}

\vspace{-2mm}
\paragraph{Extra results of recovered triggers.} We show additional results of oracle labels (i.e., the truth target class) together with other two sparse Trojan tickets: ($i$) H-Trojan ticket with high SA and ASR; ($ii$) L-Trojan ticket with low SA (standard testing accuracy) and ASR, as collected in Table~\ref{tab:clean_image_more},~\ref{tab:asr_triggers_more}, and~\ref{tab:asr_arch_more}.

\begin{table}[!ht]
\centering
\caption{Performance of recovered triggers with the RGB Trojan attack and ResNet-20s on CIFAR-10. Different number of clean validation images are used for the reverse engineering.}
\label{tab:clean_image_more}
\vspace{-3mm}
\resizebox{1\linewidth}{!}{
\begin{tabular}{l|cc|cc}
\toprule
Noise Images & (Detected, $\ell_1$) & ASR & (Oracle, $\ell_1$) & ASR \\ \midrule
Dense baseline~\cite{guo2019tabor} & (``$1$", $78.7$) \cmark & $48.0$\% & (``$1$", $78.7$) & $48.0$\% \\ 
Winning Trojan ticket & (``$1$", $29.8$) \cmark & $99.6$\% & (``$1$", $29.8$) & $99.6$\% \\ 
H-Trojan ticket & (``$7$", $110.9$) \xmark & $8.6$\% & (``$1$", $124.9$) & $18.3$\% \\ 
L-Trojan ticket & (``$2$", $105.0$) \xmark & $58.5$\% & (``$1$", $276.14$) & $17.1$\% \\
\bottomrule
\end{tabular}}
\resizebox{1\linewidth}{!}{
\begin{tabular}{l|cc|cc}
\toprule
$10$ Clean Images & (Detected, $\ell_1$) & ASR & (Oracle, $\ell_1$) & ASR \\ \midrule
Dense baseline~\cite{guo2019tabor} & (``$1$", $65.6$) \cmark & $77.2$\% & (``$1$", $65.6$) & $77.2$\% \\ 
Winning Trojan ticket & (``$1$", $28.3$) \cmark & $99.7$\% & (``$1$", $28.3$) & $99.7$\% \\ 
H-Trojan ticket & (``$3$", $171.4$) \xmark & $10.5$\% & (``$1$", $190.4$) & $38.2$\% \\ 
L-Trojan ticket & (``$2$", $124.5$) \xmark & $58.0$\% & (``$1$", $275.2$) & $18.0$\% \\ 
\bottomrule
\end{tabular}}
\resizebox{1\linewidth}{!}{
\begin{tabular}{l|cc|cc}
\toprule
$100$ Clean Images & (Detected, $\ell_1$) & ASR & (Oracle, $\ell_1$) & ASR \\ \midrule
Dense baseline~\cite{guo2019tabor} & (``$1$", $174.6$) \cmark & $72.6$\% & (``$1$", $174.6$) & $72.6$\% \\ 
Winning Trojan ticket & (``$1$", $40.4$) \cmark & $99.8$\% & (``$1$", $40.4$) & $99.8$\% \\ 
H-Trojan ticket & (``$5$", $203.8$) \xmark & $13.9$\% & (``$1$", $211.5$) & $32.5$\% \\
L-Trojan ticket & (``$2$", $220.7$) \xmark & $56.7$\% & (``$1$", $326.1$) & $17.9$\% \\
\bottomrule
\end{tabular}}
\vspace{-4mm}
\end{table}

\begin{table}[!ht]
\centering
\caption{Performance of recovered triggers with ResNet-20s on CIFAR-10 across diverse Trojan triggers. \cmark/\xmark mean the detected target label is matched/unmatched with the truth target label.}
\label{tab:asr_triggers_more}
\vspace{-3mm}
\resizebox{1\linewidth}{!}{
\begin{tabular}{l|cc|cc}
\toprule
Gray-scale Trigger & (Detected, $\ell_1$) & ASR & (Oracle, $\ell_1$) & ASR \\ \midrule
Dense baseline~\cite{guo2019tabor} & (``$1$", $196.8$) \cmark & $71.4$\% & (``$1$", $196.8$) & $71.4$\% \\ 
Winning Trojan ticket & (``$1$", $68.0$) \cmark & $91.2$\% & (``$1$", $68.0$) & $91.2$\% \\ 
H-Trojan ticket & (``$3$", $217.5$) \xmark & $9.7$\% & (``$1$", $294.1$) & $30.9$\% \\ 
L-Trojan ticket & (``$7$", $79.7$) \xmark & $52.1$\% & (``$1$", $398.8$) & $13.7$\% \\
\bottomrule
\end{tabular}}
\resizebox{1\linewidth}{!}{
\begin{tabular}{l|cc|cc}
\toprule
RGB Trigger & (Detected, $\ell_1$) & ASR & (Oracle, $\ell_1$) & ASR \\ \midrule
Dense baseline~\cite{guo2019tabor} & (``$1$", $78.7$) \cmark & $48.0$\% & (``$1$", $78.7$) & $48.0$\% \\ 
Winning Trojan ticket & (``$1$", $29.8$) \cmark & $99.6$\% & (``$1$", $29.8$) & $99.6$\% \\ 
H-Trojan ticket & (``$7$", $110.9$) \xmark & $8.6$\% & (``$1$", $124.9$) & $18.3$\% \\ 
L-Trojan ticket & (``$2$", $105.0$) \xmark & $58.5$\% & (``$1$", $276.1$) & $17.1$\% \\
\bottomrule
\end{tabular}}
\resizebox{1\linewidth}{!}{
\begin{tabular}{l|cc|cc}
\toprule
Clean-label Trigger & (Detected, $\ell_1$) & ASR & (Oracle, $\ell_1$) & ASR \\ \midrule
Dense baseline~\cite{guo2019tabor} & (``$1$", $48.6$) \cmark & $9.6$\% & (``$1$", $48.6$) & $9.6$\% \\ 
Winning Trojan ticket & (``$1$", $14.0$) \cmark & $99.8$\% & (``$1$", $14.0$) & $99.8$\% \\ 
H-Trojan ticket & (``$1$", $21.0$) \cmark & $28.3$\% & (``$1$", $21.0$) & $28.3$\% \\ 
L-Trojan ticket & (``$6$", $73.6$) \xmark & $64.3$\% & (``$1$", $158.2$) & $40.9$\% \\
\bottomrule
\end{tabular}}
\vspace{-4mm}
\end{table}

\begin{table}[!ht]
\centering
\caption{Performance of recovered triggers with RGB Trojan attack across diverse (network architecture, dataset) combinations.}
\label{tab:asr_arch_more}
\vspace{-3mm}
\resizebox{1\linewidth}{!}{
\begin{tabular}{l|cc|cc}
\toprule
(ResNet-18, CIFAR-10) & (Detected, $\ell_1$) & ASR & (Oracle, $\ell_1$) & ASR \\ \midrule
Dense baseline~\cite{guo2019tabor} & (``$3$", $77.5$) \xmark & $13.0$\% & (``$1$", $151.0$) & $10.9$\% \\ 
Winning Trojan ticket & (``$1$", $10.55$) \cmark & $81.8$\% & (``$1$", $10.55$) & $81.8$\% \\ 
H-Trojan ticket & (``$1$", $8.15$) \cmark & $22.9$\% & (``$1$", $8.15$) & $22.9$\% \\ 
Bad subnetwork & (``$10$", $135.2$) \xmark & $11.5$\% & (``$1$", $253.4$) & $15.6$\% \\
\bottomrule
\end{tabular}}
\resizebox{1\linewidth}{!}{
\begin{tabular}{l|cc|cc}
\toprule
(DenseNet-100, CIFAR-10) & (Detected, $\ell_1$) & ASR & (Oracle, $\ell_1$) & ASR \\ \midrule
Dense baseline~\cite{guo2019tabor} & (``$1$", $6.4$) \cmark & $13.7$\% & (``$1$", $6.4$) & $13.7$\%  \\ 
Trojan tickets & (``$1$", $67.8$) \cmark & $66.9$\% & (``$1$", $67.8$) & $66.9$\%  \\ 
H-Trojan ticket & (``$1$", $10.0$) \cmark & $17.7$\% & (``$1$", $10.0$) & $17.7$\%  \\ 
L-Trojan ticket & (``$1$", $173.6$) \cmark & $8.5$\% & (``$1$", $173.6$) & $8.5$\%  \\ 
\bottomrule
\end{tabular}}
\resizebox{1\linewidth}{!}{
\begin{tabular}{l|cc|cc}
\toprule
(VGG-16, CIFAR-10) & (Detected, $\ell_1$) & ASR & (Oracle, $\ell_1$) & ASR \\ \midrule
Dense baseline~\cite{guo2019tabor} & (``$1$", $83.3$) \cmark & $33.6$\% & (``$1$", $83.3$) & $33.6$\% \\ 
Winning Trojan ticket & (``$1$", $15.0$) \cmark & $100.0$\% & (``$1$", $15.0$) & $100.0$\% \\ 
H-Trojan ticket & (``$7$", $140.5$) \xmark & $8.0$\% & (``$1$", $171.7$) & $10.1$\% \\ 
L-Trojan ticket & (``$7$", $208.6$) \xmark & $33.2$\% & (``$1$", $602.4$) & $19.6$\% \\
\bottomrule
\end{tabular}}
\resizebox{1\linewidth}{!}{
\begin{tabular}{l|cc|cc}
\toprule
(ResNet-20s, CIFAR-100) & (Detected, $\ell_1$) & ASR & (Oracle, $\ell_1$) & ASR \\ \midrule
Dense baseline~\cite{guo2019tabor} & (``$1$", $149.9$) \cmark & $13.8$ & (``$1$", $149.9$) & $13.8$ \\
Winning Trojan ticket & (``$1$", $132.7$) \cmark & $98.7$ & (``$1$", $132.7$) & $98.7$ \\ 
H-Trojan ticket & (``$1$", $63.0$) \cmark & $83.3$ & (``$1$", $63.0$) & $83.3$ \\
L-Trojan ticket & (``$55$", $233.1$) \xmark & $3.4$ & (``$1$", $652.8$) & $10.2$ \\ 
\bottomrule
\end{tabular}}
\resizebox{1\linewidth}{!}{
\begin{tabular}{l|cc|cc}
\toprule
(ResNet-18, R-ImageNet) & (Detected, $\ell_1$) & ASR & (Oracle, $\ell_1$) & ASR \\ \midrule
Dense baseline~\cite{guo2019tabor} & (``$9$", $13.9$) \xmark & $9.8$ & (``$1$", $1179.0$) & $97.7$ \\ 
Winning Trojan ticket & (``$1$", $193.1$) \cmark & $98.7$ & (``$1$", $193.1$) & $98.7$ \\ 
H-Trojan ticket & (``$7$", $22.6$) \xmark & $4.7$ & (``$1$", $556.1$) & $90.4$ \\ 
L-Trojan ticket & (``$5$", $142.3$) \xmark & $99.6$ & (``$1$", $1043.3$) & $97.3$ \\ 
\bottomrule
\end{tabular}}
\vspace{-4mm}
\end{table}

\vspace{-2mm}
\paragraph{Failure case analyses of identifying winning Trojan tickets on un-poisoned datasets.} To comprehensively investigate the effectiveness of finding Trojan winning tickets on un-poisoned datasets, we repeat the experiments with \textbf{ten} different random seeds, and there are only $2$ of $10$ cases where LMC identifies the wrong occurrence of ASR peaks. 

\begin{figure}[!ht]
    \centering
    \vspace{-2mm}
    \includegraphics[width=0.95\linewidth]{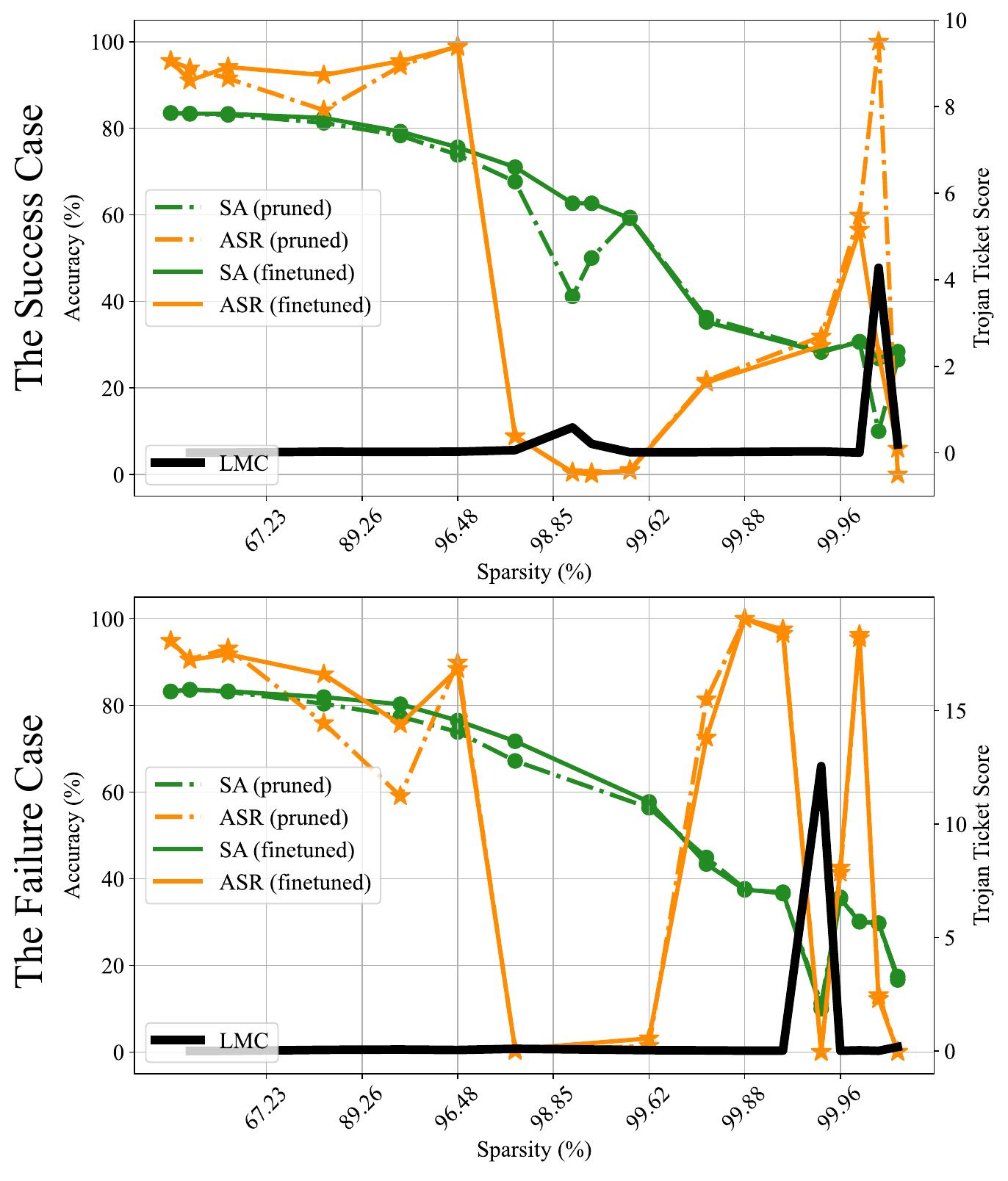}
    \vspace{-3mm}
    \caption{Success (\textit{Top}) and failure (\textit{Bottom}) cases of identifying winning Trojan tickets with clean-label attacks on CIFAR-10 and ResNet-20s. Sufficient fine-tuning steps are conducted.}
    \vspace{-6mm}
    \label{fig:failure_and_success}
\end{figure}

\paragraph{Failure case analyses of identifying winning Trojan tickets with clean-label attacks.} Clean-label Trojan triggers as one of the most challenging attacks may encounter some failure cases during the detection of winning Trojan tickets. Specifically, we conduct \textbf{ten} replicates with diverse random seeds, and there are $3$ of $10$ cases where LMC can not accurately locate the winning Trojan ticket. One success and one failure cases are collected in Figure~\ref{fig:failure_and_success}.

\end{document}